\documentclass{article}

\usepackage{amsmath}
\usepackage{mathtools}
\usepackage{amsfonts}

\usepackage{microtype}
\usepackage{graphicx}
\usepackage{subfigure}
\usepackage{booktabs} 

\usepackage{caption}
\usepackage{pdfpages}
\usepackage{thmtools}
\usepackage{thm-restate}

\usepackage{hyperref}

\usepackage[accepted]{icml2019}

\icmltitlerunning{Boosting Trust Region Policy Optimization with Normalizing Flows Policy}

\begin{document}

\twocolumn[
\icmltitle{Boosting Trust Region Policy Optimization with Normalizing Flows Policy}




\begin{icmlauthorlist}
\icmlauthor{Yunhao Tang}{columbia}
\icmlauthor{Shipra Agrawal}{columbia}
\end{icmlauthorlist}

\icmlaffiliation{columbia}{Columbia University, New York, NY, USA}

\icmlcorrespondingauthor{Yunhao Tang}{yt2541@columbia.edu}

\icmlkeywords{Machine Learning, ICML}

\vskip 0.3in
]



\printAffiliationsAndNotice{}  

\begin{abstract}
We propose to improve trust region policy search with normalizing flows policy. We illustrate that when the trust region is constructed by KL divergence constraints, normalizing flows policy generates samples far from the 'center' of the previous policy iterate, which potentially enables better exploration and helps avoid bad local optima. Through extensive comparisons, we show that the normalizing flows policy significantly improves upon baseline architectures especially on high-dimensional tasks with complex dynamics.
\end{abstract}

\section{Introduction}
In on-policy optimization, vanilla policy gradient algorithms suffer from occasional updates with large step size, which lead to collecting bad samples that the policy cannot recover from \citep{schulman2015}. Motivated to overcome such instability, Trust Region Policy Optimization (TRPO) \citep{schulman2015} constraints the KL divergence between consecutive policies to achieve much more stable updates. However, with factorized Gaussian policy, such KL divergence constraint can put a very stringent restriction on the new policy iterate, making it hard to bypass locally optimal solutions and slowing down the learning process.

Can we improve the learning process of trust region policy search by using a more expressive policy class? Intuitively, a more expressive policy class has more capacity to represent complex distributions and as a result, the KL constraint may not impose a very strict restriction on the sample space. Though prior works \citep{tuomas2017,tuomas2018,tuomas2018b} have proposed to use expressive generative models as policies, their focus is on off-policy learning. In this work, we show how expressive distributions, in particular normalizing flows \citep{rezende2015,dinh2017} can be combined with on-policy learning and boost the performance of trust region policy optimization.


The structure of our paper is as follows. In Section 2 and 3, we provide backgrounds and related work. In Section 4, we introduce normalizing flows for control and analyze why KL constraint may not impose a restrictive constraint on the sampled action space. With illustrative examples, we show that normalizing flows policy can learn policies with correlated actions and multi-modal policies, which allows for potentially more efficient exploration. In Section 5, we show by comprehensive experiments that normalizing flows significantly outperforms baseline policy classes when combined with trust region policy search algorithms.

\section{Background}
\subsection{Markov Decision Process}
In the standard formulation of Markov Decision Process (MDP), at time step $t\geq0$, an agent is in state $s_t \in \mathcal{S}$, takes an action $a_t \in \mathcal{A}$, receives an instant reward $r_t = r(s_t,a_t) \in \mathbb{R}$ and transitions to a next state $s_{t+1} \sim p(\cdot|s_t,a_t) \in \mathcal{S}$. Let $\pi : \mathcal{S} \mapsto P(\mathcal{A})$ be a policy, where $P(\mathcal{A})$ is the set of distributions over the action space $\mathcal{A}$. The discounted cumulative reward under policy $\pi$ is $J(\pi) = \mathbb{E}_{\pi}\big[\sum_{t=0}^\infty \gamma^t r_t\big]$, where $\gamma \in [0,1)$ is a discount factor. The objective of RL is to search for a policy $\pi$ that achieves the maximum cumulative reward $\pi^\ast = \arg\max_\pi J(\pi)$. For convenience, under policy $\pi$ we define action value function $Q^\pi(s,a) = \mathbb{E}_{\pi}\big[J(\pi)|s_0=s,a_0=a\big]$ and value function $V^\pi(s) = \mathbb{E}_\pi\big[J(\pi)|s_0=s,a_0\sim \pi(\cdot|s_0)\big]$. We also define the advantage function $A^\pi(s,a) = Q^\pi(s,a) - V^\pi(s)$.

\subsection{Policy Optimization}
One way to approximately find $\pi^\ast$ is through direct policy search within a given policy class $\pi_\theta,\theta\in\Theta$ where $\Theta$ is the parameter space for the policy parameter. We can update the paramter $\theta$ with policy gradient ascent, by computing
$\nabla_\theta J(\pi_\theta) = \mathbb{E}_{\pi_\theta}\big[\sum_{t=0}^\infty A^{\pi_\theta}(s_t,a_t) \nabla_\theta \log \pi_\theta(a_t|s_t)\big]$, then updating $\theta_{\text{new}} \leftarrow \theta + \alpha \nabla_\theta J(\pi_\theta)$ with some learning rate $\alpha > 0$. Alternatively, the update can be formulated by first considering a trust region optimization problem
\begin{align}
\max_{\theta_{\text{new}}}\ \mathbb{E}_{\pi_\theta}\big[&\frac{\pi_{\theta_\text{new}}(a_t|s_t)}{\pi_\theta(a_t|s_t)} A^{\pi_\theta}(s_t,a_t)\big], \nonumber \\
&||\theta_{\text{new}} - \theta||_2 \leq \epsilon,
\label{eq:trustregionpg}
\end{align}
for some $\epsilon > 0$. If we do a linear approximation of the objective in (\ref{eq:trustregionpg}), $\mathbb{E}_{\pi_\theta}\big[\frac{\pi_{\theta_\text{new}}(a_t|s_t)}{\pi_\theta(a_t|s_t)} A^{\pi_\theta}(s_t,a_t)\big] \approx \nabla_\theta J(\pi_\theta)^T (\theta_{\text{new}} - \theta)$, we recover the policy gradient update by properly choosing $\epsilon$ given $\alpha$.

\subsection{Trust Region Policy Optimization}
Trust Region Policy Optimization (TRPO) \citep{schulman2015} applies information theoretic constraints instead of Euclidean constraints (as in (\ref{eq:trustregionpg})) between $\theta_{\text{new}}$ and $\theta$ to better capture the geometry on the parameter space induced by the underlying distributions. In particular, consider the following trust region formulation
\begin{align}
\max_{\theta_{\text{new}}}\  &\mathbb{E}_{\pi_\theta}\big[\frac{\pi_{\theta_\text{new}}(a_t|s_t)}{\pi_\theta(a_t|s_t)} A^{\pi_\theta}(s_t,a_t)\big], \nonumber \\
&\mathbb{E}_{s} \big[\mathbb{KL}[\pi_\theta(\cdot|s) || \pi_{\theta_{\text{new}}}(\cdot|s)]\big] \leq \epsilon,
\label{eq:trpo}
\end{align}
where $\mathbb{E}_s\big[\cdot\big]$ is w.r.t. the state visitation distribution induced by $\pi_\theta$. The trust region enforced by the KL divergence entails that the update according to (\ref{eq:trpo}) optimizes a lower bound of $J(\pi_\theta)$, so as to avoid accidentally taking large steps that irreversibly degrade the policy performance during training as with vanilla policy gradient (\ref{eq:trustregionpg}) \citep{schulman2015}. To make the algorithm practical, the trust region constraint is approximated by a second order expansion $\mathbb{E}_{s}\big[\mathbb{KL}[\pi_\theta(\cdot|s) || \pi_{\theta_{\text{new}}}(\cdot|s)]\big] \approx (\theta_{\text{new}} - \theta)^T \hat{H}  (\theta_{\text{new}} - \theta) \leq \epsilon$ where $\hat{H} = \frac{\partial^2}{\partial \theta^2} \mathbb{E}_{\pi_\theta}\big[\mathbb{KL}[\pi_\theta(\cdot|s) || \pi_{\theta_{\text{new}}}(\cdot|s)]\big] $ is the expected Fisher information matrix. If we also linearly approximate the objective, the trust region formulation turns into a quadratic programming
\begin{align}
\max_{\theta_{\text{new}}}  \nabla_\theta J(\pi_\theta)^T (\theta_{\text{new}} - \theta), \nonumber \\
(\theta_{\text{new}} - \theta)^T \hat{H}  (\theta_{\text{new}} - \theta) \leq \epsilon.
\label{eq:approxtrpo}
\end{align}
The optimal solution to (\ref{eq:approxtrpo}) is $\propto \hat{H}^{-1} \nabla_\theta J(\pi_\theta)$. In cases where $\pi_\theta$ is parameterized by a neural network with a large number of parameters, $\hat{H}^{-1}$ is formidable to compute. Instead, \citet{schulman2015} propose to approximate $\hat{H}^{-1} \nabla_\theta J(\pi_\theta)$ by conjugate gradient (CG) descent \citep{wright1999numerical} since it only requires relatively cheap Hessian-vector products. Given the approximated update direction $\hat{g} \approx \hat{H}^{-1} \nabla_\theta J(\pi_\theta) $ obtained from CG, the KL constraint is enforced by setting $\Delta \theta = \sqrt{\frac{\epsilon}{\hat{g}^T \hat{H} \hat{g}}} \hat{g}$. Finally a line search is carried out to determine a scaler $s$ by enforcing the exact KL constraint $\mathbb{E}_{\pi_\theta}\big[\mathbb{KL}[\pi_{\theta + s \Delta \theta} || \pi_\theta ]\big] \leq \epsilon$ and finally $\theta_{\text{new}} \leftarrow \theta + s\Delta \theta$.

\paragraph{ACKTR.} ACKTR \citet{wu2017scalable} propose to replace the above CG descent of TRPO by Kronecker-factored approximation \citep{martens2015optimizing} when computing the inverse of Fisher information matrix $\hat{H}^{-1}$. This approximation is more stable than CG descent and yields performance gain over conventional TRPO.

\subsection{Normalizing flows}
Normalizing flows  \citep{rezende2015,dinh2017} have been applied in variational inference and probabilistic modeling to represent complex distributions. In general, consider transforming a source noise $\epsilon \sim \rho_0(\cdot)$ by a series of invertible nonlinear functions $g_{\theta_i}(\cdot),1\leq i\leq K$ each with parameter $\theta_i$, to output a target sample $x$,
\begin{align}
x = g_{\theta_K} \circ g_{\theta_{K-1}} \circ ... \circ g_{\theta_2} \circ g_{\theta_1} (\epsilon).
\label{eq:normflow}
\end{align}
Let $\Sigma_i$ be the inverse of the Jacobian matrix of $g_\theta(\cdot)$, then the log density of $x$ is computed by the change of variables formula,
\begin{align}
\log p(x) = \log p(\epsilon) + \sum_{i=1}^K \log \text{det}(\Sigma_i).
\label{eq:chainrule}
\end{align}
The distribution of $x$ is determined by the noise $\epsilon$ and the transformations $g_{\theta_i}(\cdot)$. When the transformations $g_{\theta_i}(\cdot)$ are very complex (e.g. $g_{\theta_i}(\cdot)$ are neural networks) we expect $p(x)$ to be highly expressive as well. However, for a general invertible transformation $g_{\theta_i}(\cdot)$, computing the determinant $\text{det}(\Sigma_i)$ is expensive. In this work, we follow the architecture of \citep{dinh2015} to ensure that $\text{det}(\Sigma_i)$ is computed cheaply while $g_{\theta_i}(\cdot)$ are expressive. We leave all details to Appendix B. We will henceforth also address the normalizing flows policy as the NF policy.

\section{Related Work}
\paragraph{On-Policy Optimization.} In on-policy optimization, vanilla policy gradient updates are generally unstable \citep{schulman2015}. Natural policy gradient \citep{kakade2002natural} applies natural gradient for policy updates, which accounts for the information geometry induced by the policy and enables more stable updates. More recently, Trust region policy optimization \citep{schulman2015} derives a scalable trust region policy search algorithm based on the lower bound formulation of \citep{kakade2002approximately} and achieves promising results on simulated locomotion tasks. To further improve the performance of TRPO, ACKTR \citep{wu2017scalable} applies Kronecker-factored approximation \citep{martens2015optimizing} to construct the updates. Orthogonal to prior works, we aim to improve TRPO with a more expressive policy class, and we show significant improvements on both TRPO and ACKTR. 

\paragraph{Policy Classes.} Several recent prior works have proposed to boost RL algorithms with expressive policy classes. Thus far, expressive policy classes have shown improvement over baselines in off-policy learning: Soft Q-learning (SQL) \citep{tuomas2017} takes an implicit generative model as the policy and trains the policy by Stein variational gradients \citep{liu2016}; \citet{tang2018implicit} applies an implicit policy along with a discriminator to compute entropy regularized gradients. Latent space policy \citep{tuomas2018b} applies normalizing flows as the policy and displays promising results on hierarchical tasks; Soft Actor Critic (SAC) \citep{tuomas2018} applies a mixture of Gaussian as the policy. However, aformentioned prior works do not disentangle the architectures from the algorithms, it is hence not clear whether the gains come from an expressive policy class or novel algorithmic procedures. In this work, we fix the trust region search algorithms and study the net effect of expressive policy classes.

For on-policy optimization, Gaussian policy is the default baseline \citep{schulman2015,schulman2017}. Recently, \citep{chou2017improving} propose Beta distribution as an alternative to Gaussian and show improvements on TRPO. We make a full comparison and show that expressive distributions achieves more consistent and stable gains than such bounded distributions for TRPO.

\paragraph{Normalizing flows.} By construction, normalizing flows stacks layers of invertible transformations to map a source noise into target samples \citep{dinh2015,dinh2017,rezende2015}. Normalizing flows retains tractable densities while being very expressive, and is widely applied in probabilistic generative modeling and variational inference \citep{rezende2015,louizos2017multiplicative}. Complement to prior works, we show that normalizing flows can significantly boost the performance of TRPO/ACKTR. We limit our attention to the architectures of \citep{dinh2017} while more recent flows structure might offer additional performance gains \citep{kingma2018glow}.

\section{Normalizing flows Policy for On-Policy Optimization}

\subsection{Normalizing flows for control}
We now construct a stochastic policy based on normalizing flows. By design, we require the source noise $\epsilon$ to have the same dimension as the action $a$. Recall that the normalizing flows distribution is implicitly defined by a sequence of invertible transformation (\ref{eq:normflow}). To define a proper policy $\pi(a|s)$, we first embed state $s$ by another neural network $L_{\theta_s}(\cdot)$ with parameter $\theta_s$ and output a state vector $ L_{\theta_s}(s)$ with the same dimension as $\epsilon$. We can then insert the state vector between any two layers of (\ref{eq:normflow}) to make the distribution conditional on state $s$. In our implementation, we insert the state vector after the first transformation (for clarity of the presentation, we leave all details of the architectures of $g_{\theta_i}$  and $L_{\theta_s}$ in Appendix B).
\begin{align}
a = g_{\theta_K} \circ g_{\theta_{K-1}} \circ ... \circ g_{\theta_2} \circ (L_{\theta_s}(s) + g_{\theta_1} (\epsilon)).
\label{eq:statenormflow}
\end{align}
Though the additive form of $L_{\theta_s}(s)$ and $g_{\theta_1}(\epsilon)$ may in theory limit the capacity of the model, in experiments below we show that the resulting policy is still very expressive. For simplicity, we denote the above transformation (\ref{eq:statenormflow}) as $a = f_\theta(s,\epsilon)$ with parameter $\theta = \{\theta_s,\theta_i,1\leq i\leq K\}$. It is obvious that the transformation $a = f_\theta(s,\epsilon)$ is still invertible between $a$ and $\epsilon$, which is critical for computing $\log \pi_\theta(a|s)$ according to the change of variables formula (\ref{eq:chainrule}). This architecture builds complex action distributions with explicit probability density $\pi_\theta(\cdot|s)$, and hence entails training using score function gradient estimators.

In implementations, it is necessary to compute gradients of the entropy $\nabla_\theta \mathbb{H}\big[\pi_\theta(\cdot|s)\big]$, either for computing Hessian vector product for CG \citep{schulman2015} or for entropy regularization \citep{schulman2015,schulman2017,mnih2016}. For normalizing flows there is no analytic form for entropy, instead we can use $N$ samples to estimate entropy by re-parameterization,
 $\mathbb{H}\big[\pi_\theta(\cdot|s)\big] = \mathbb{E}_{a \sim \pi_\theta(\cdot|s)}\big[-\log \pi_\theta(a|s)\big] = \mathbb{E}_{\epsilon \sim \rho_0(\cdot)}\big[-\log \pi_\theta(f_\theta(s,\epsilon)|s)\big] \approx \frac{1}{N}\sum_{i=1}^N -\log \pi_\theta(f_\theta(s,\epsilon_i)|s)$. The gradient of the entropy can be easily estimated with samples and implemented using back-propagation
 $\nabla_\theta \mathbb{H}\big[\pi_\theta(\cdot|s)\big]  \approx \frac{1}{N}\sum_{i=1}^N -\nabla_\theta \log \pi_\theta(f_\theta(s,\epsilon_i)|s)\big]$.
 

\subsection{Understanding Normalizing flows Policy for Control}
In the following we illustrate the empirical properties of NF policy on toy examples.

\begin{figure}[h]
\centering
\subfigure[KL ball: Contours]{\includegraphics[width=.39\linewidth]{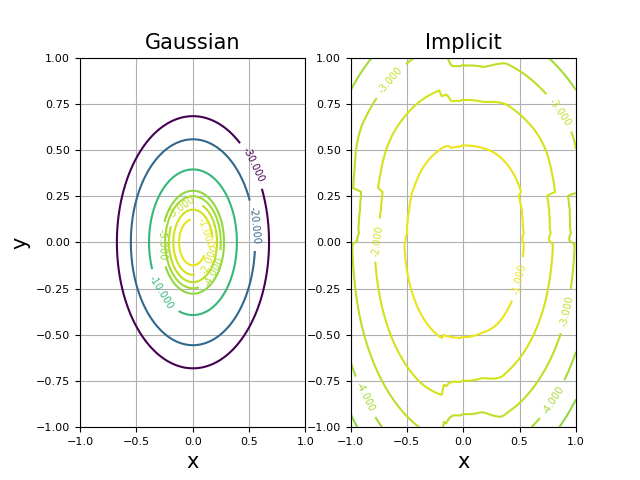}}
\subfigure[KL ball: Samples]{\includegraphics[width=.44\linewidth]{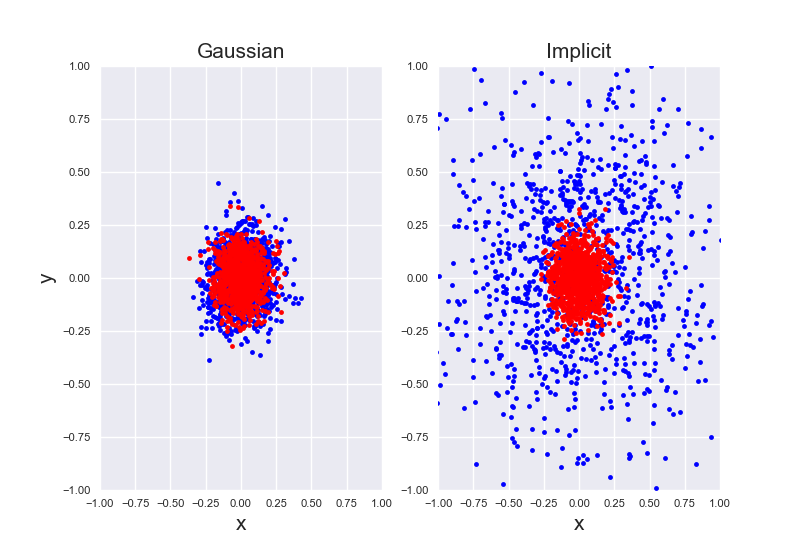}}
\caption{\small{Analyzing normalizing flows vs. Gaussian: Consider a 2D Gaussian distribution with zero mean and factorized variance $\sigma^2 = 0.1^2$. Samples from the Gaussian form an empirical distribution $\hat{\pi}_o$ (red dots in (b)) and define the KL ball $\mathcal{B}(\hat{\pi}_o,\epsilon) = \{\pi: \mathbb{KL}[\hat{\pi}_o || \pi]  \leq \epsilon\}$ centered at $\hat{\pi}_o$. We then find a NF distribution and a Gaussian distribution at the boundary of $\mathcal{B}(\hat{\pi},0.01)$ such that the constraint is tight. (a) Contour of log probability of a normalizing flows distribution (right) vs. Gaussian distribution (left); (b) Samples (blue dots) generated from NF distribution (right) and Gaussian distribution (left).}}
\label{figure:analysis}
\end{figure}

\paragraph{Generating Samples under KL Constraints.} We analyze the properties of NF policy vs. Gaussian policy under the KL constraints of TRPO. As a toy example, assume we have a factorized Gaussian in $\mathbb{R}^2$ with zero mean and diagonal covariance $\mathbb{I}\cdot \sigma^2$ where $\sigma^2 = 0.1^2$. Let $\hat{\pi}_o$ be the empirical distribution formed by samples drawn from this Gaussian. We can define a KL ball centered on $\hat{\pi}_o$ as all distributions such that a KL constraint is satisfied $\mathcal{B}(\hat{\pi}_o,\epsilon) = \{\pi: \mathbb{KL}[\hat{\pi}_o || \pi]  \leq \epsilon\}$. We study a typical normalizing flows distribution and a factorized Gaussian distribution on the boundary of such a KL ball (such that $\mathbb{KL}[\hat{\pi}_o|| \pi] = \epsilon$). We obtain such distributions by randomly initializing the distribution parameters and then running gradient updates until $\mathbb{KL}[\hat{\pi}_o || \pi]  \approx \epsilon$. In Figure \ref{figure:analysis} (a) we show the log probability contour of such a factorized Gaussian vs. normalizing flows, and in (b) we show their samples (blue are samples from the distributions on the boundary of the KL ball and red are empirical samples of $\hat{\pi}_o$). As seen from both plots, though both distributions satisfy the KL constraints, normalizing flows distribution has much larger variance than the factorized Gaussian, which also leads to a much larger effective support.




Such sample space properties have practical implications. For a factorized Gaussian distribution, enforcing KL constraints does not allow the new distribution to generate samples that are too far from the 'center' of the old distribution. On the other hand, for a normalizing flows distribution, the KL constraint does not hinder the new distribution to have a very distinct support from the reference distribution (or the old policy iterate), hence allowing for more efficient exploration that bypasses local optimal in practice.

\begin{figure}[h]
\centering
\subfigure[Correlated Actions]{\includegraphics[width=.45\linewidth]{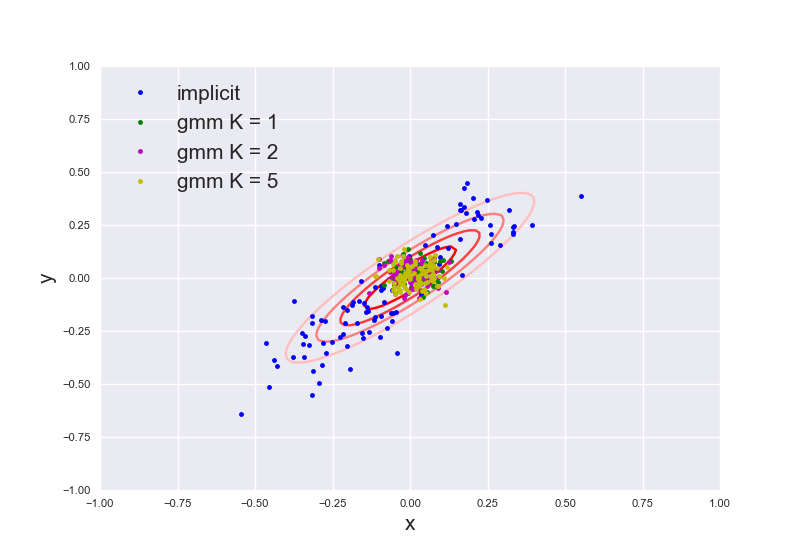}}
\subfigure[Bimodal Rewards]{\includegraphics[width=.45\linewidth]{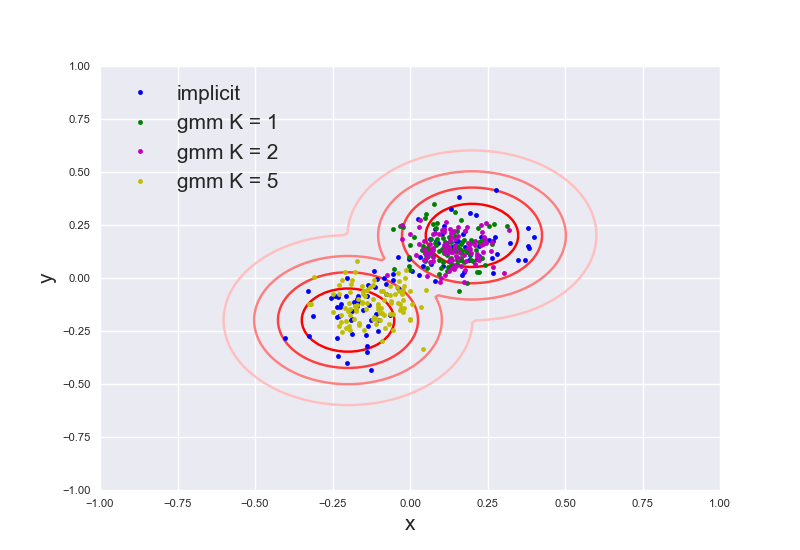}}
\subfigure[Ant Trajectories]{\includegraphics[width=.45\linewidth]{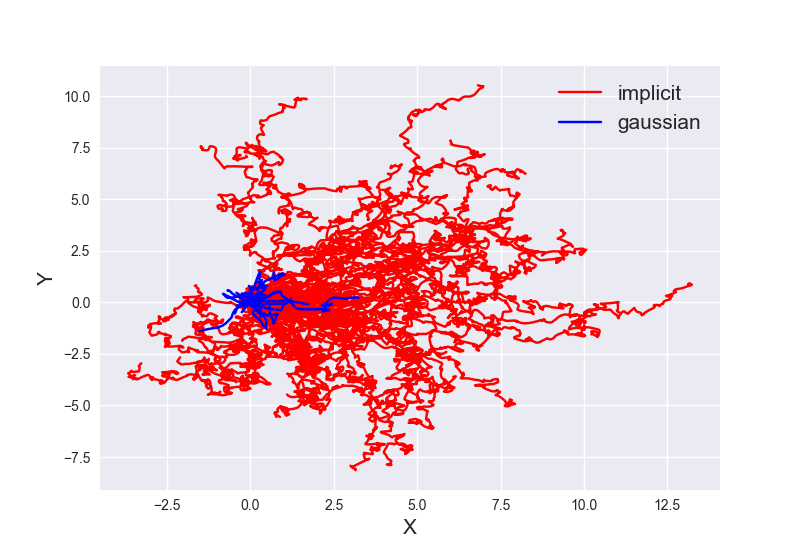}}
\subfigure[Ant Illustration]{\includegraphics[width=.4\linewidth]{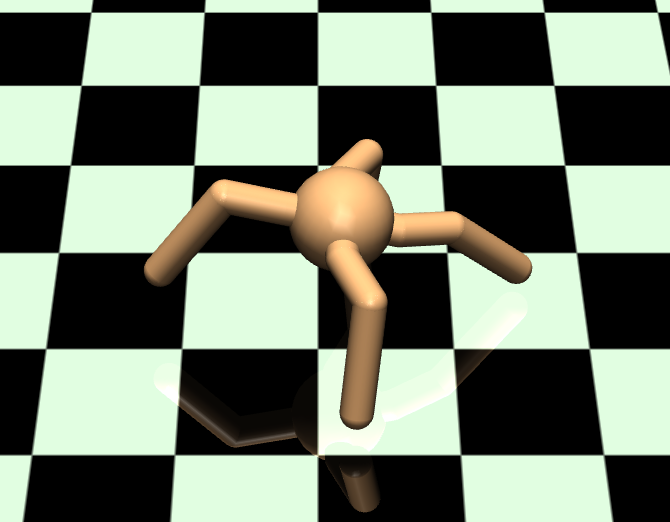}}
\caption{\small{Expressiveness of NF policy: (a) Bandit problem with reward $r(a) = -a^T \Sigma^{-1} a$. The maximum entropy optimal policy is a Gaussian distribution with $\Sigma$ as its covariance (red contours). The NF policy (blue) can capture such covariance while Gaussian cannot (green). (b) Bandit problem with multi-modal reward (red contours the reward landscape). normalizing flows policy can capture the multi-modality (blue) while Gaussian cannot (green). (c) Trajectories of Ant robot. The trajectories of Gaussian policy center at the initial position (the origin $[0.0,0.0]$), while trajectories of NF policy are much more widespread. (d) Illustration of the Ant locomotion task.}}

\label{figure:expressive}
\end{figure}

\paragraph{Expressiveness of Normalizing flows Policy.}
We illustrate two potential strengths of the NF policy: learning correlated actions and learning multi-modal policy. First consider a 2D bandit problem where the action $a \in [-1,1]^2$ and $r(a) = -a^T \Sigma^{-1} a$ for some positive semidefinite matrix $\Sigma$. In the context of conventional RL objective $J(\pi)$, the optimal policy is deterministic $\pi^\ast = [0,0]^T$. However, in maximum entropy RL \citep{tuomas2017,tuomas2018} where the objective is $J(\pi) + c\mathbb{H}\big[\pi\big]$, the optimal policy is $\pi_{\text{ent}}^\ast \propto \exp(\frac{r(a)}{c})$, a Gaussian with $\frac{\Sigma}{c}$ as the covariance matrix (red curves show the density contours). In Figure \ref{figure:expressive} (a), we show the samples generated by various trained policies to see whether they manage to learn the correlations between actions in the maximum entropy policy $\pi_{\text{ent}}^\ast$. We find that the factorized Gaussian policy cannot capture the correlations due to the factorization. Though Gaussian mixtures models (GMM) with $K\geq 2$ components are more expressive than factorized Gaussian, all the modes tend to collapse to the same location and suffer the same issue as factorized Gaussian. On the other hand, the NF policy is much more flexible and can fairly accurately capture the correlation structure of  $\pi_{\text{ent}}^\ast$.

To illustrate multi-modality, consider again a 2D bandit problem (Figure \ref{figure:expressive} (b)) with reward $r(a) = \max_{i\in I} \{(a - \mu_i)^T \Lambda_i^{-} (a-\mu_i)\}$ where $\Lambda_i,i\in I$ are diagonal matrices and $\mu_i,i\in I$ are modes of the reward landscape. In our example we set $|I| = 2$ two modes and the reward contours are plotted as red curves. Notice that GMM with varying $K$ can still easily collapse to one of the two modes while the NF policy generates samples that cover both modes. 

To summarize the above two cases, since the maximum entropy objective $J(\pi) + c \mathbb{H}[\pi] =  - \mathbb{KL}[\pi || \pi_{\text{ent}}^\ast ]$, the policy search problem is equivalent to a variational inference problem where the variational distribution is $\pi$ and the target distribution is $\pi_{\text{ent}}^\ast \propto \exp(\frac{r(a)}{c})$. Since NF policy is a more expressive class of distribution than GMM and factorized Gaussian, we also expect the approximation to the target distribution to be much better \citep{rezende2015}. This \emph{control as inference} perspective provides partial justification as to why an expressive policy such as NF policy achieves better practical performance. We review such a view in Appendix E.

The properties of NF policy illustrated above potentially allow for better exploration during training and help bypass bad local optima. For a more realistic example, we illustrate such benefits with the locomotion task of Ant robot (Figure \ref{figure:expressive} (d))  \citep{brockman2016}. In Figure \ref{figure:expressive} (c) we show the robot's 2D center-of-mass trajectories generated by NF policy (red) vs. Gaussian policy (blue) after training for $2\cdot 10^6$ time steps. We observe that the trajectories by NF policy are much more widespread, while trajectories of Gaussian policy are quite concentrated at the initial position (the origin $[0.0,0.0]$). Behaviorally, the Gaussian policy gets the robot to jump forward quickly, which achieves high immediate rewards but terminates the episode prematurely (due to a termination condition of the task). On the other hand, the NF policy bypasses such locally optimal behavior by getting the robot to move forward in a fairly slow but steady manner, even occasionally move in the opposite direction to the high reward region (Task details in Appendix D).

\section{Experiments}
In experiments we aim to address the following questions: \textbf{(a)} Do NF policies outperform simple policies (e.g. factorized Gaussian baseline) as well as recent prior works (e.g. \citep{chou2017improving}) with trust region search algorithms on benchmark tasks, and especially on high-dimensional tasks with complex dynamics? \textbf{(b)} How sensitive are NF policies to hyper-parameters compared to Gaussian policies? 

To address \textbf{(a)}, we carry out comparison in three parts: (1) We implement and compare NF policy with Gaussian mixtures model (GMM) policy and factorized Gaussian policy, on a comprehensive suite of tasks in OpenAI gym MuJoCo \citep{brockman2016,todorov2008}, rllab \citep{duanxi2016}, Roboschool Humanoid \citep{schulman2017proximal} and Box2D \cite{brockman2016} illustrated in Appendix F; (2) We implement and compare NF with straightforward architectural alternatives and prior work \citep{chou2017improving} on tasks with complex dynamics; (3) Lastly but importantly, we compare NF with Gaussian policy results directly reported in prior works \citep{schulman2017proximal,wu2017scalable,chou2017improving,tuomas2018}. For (1) we show results for both TRPO and ACKTR, and for (2)(3) only for TRPO, all presented in Section 5.1. Here, we note that (3) is critical because the performance of the same algorithm across papers can be very different \citep{henderson2017deep}, and here we aim to show that our proposed method achieves significant gains consistently over results reported in prior works. To address \textbf{(b)}, we randomly sample hyper-parameters for both NF policy and Gaussian policy, and compare their quantiles in Section 5.2. 




\paragraph{Implementation Details.} As we aim to study the net effect of an expressive policy on trust region policy search, we make minimal modification to the original TRPO/ACKTR algorithms based on OpenAI baselines \citep{baselines}. For both algorithms, the policy entropy $\mathbb{H}\big[\pi_\theta(\cdot|s)\big]$ is analytically computed whenever possible, and is estimated by samples when $\pi_\theta$ is GMM for $K\geq 2$ or normalizing flows. The KL divergence is approximated by samples instead of analytically computed in a similar way. We leave all hyper-parameter settings in Appendix A. 

\subsection{Locomotion Benchmarks}

\begin{figure*}[t]
\centering
\subfigure[\textbf{Reacher}]{\includegraphics[width=.23\linewidth]{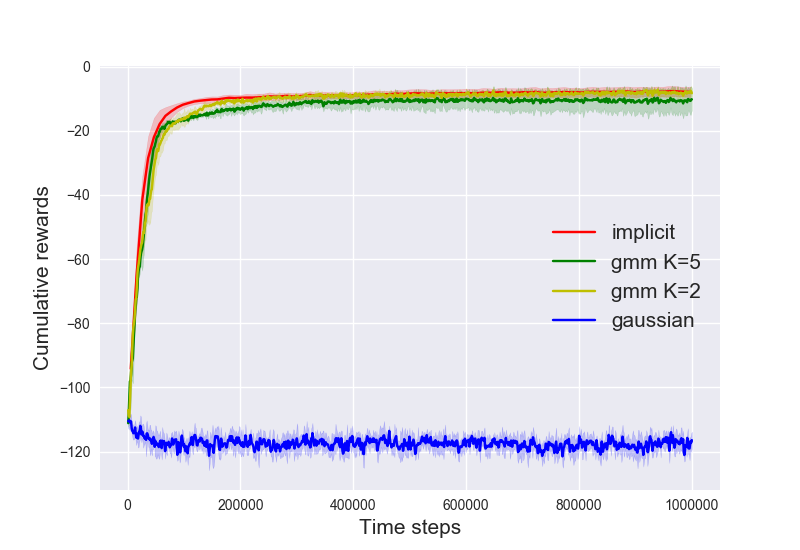}}
\subfigure[\textbf{Swimmer}]{\includegraphics[width=.23\linewidth]{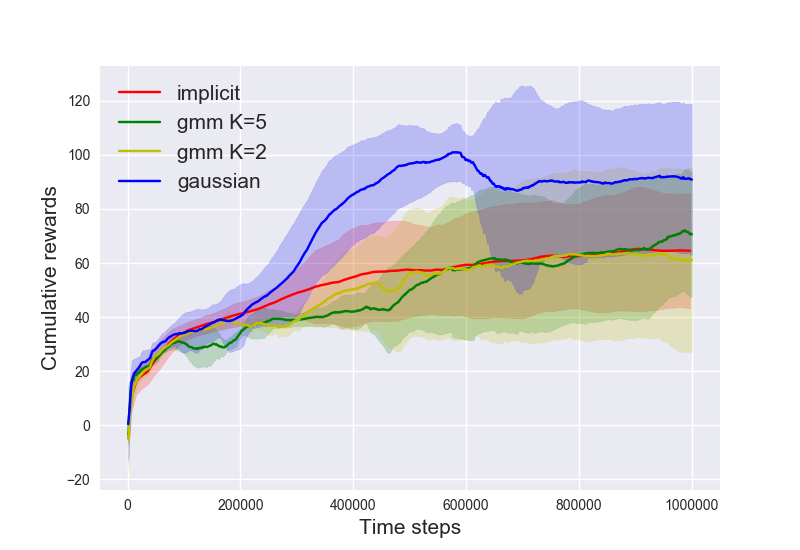}}
\subfigure[\textbf{Inverted Pendulum}]{\includegraphics[width=.23\linewidth]{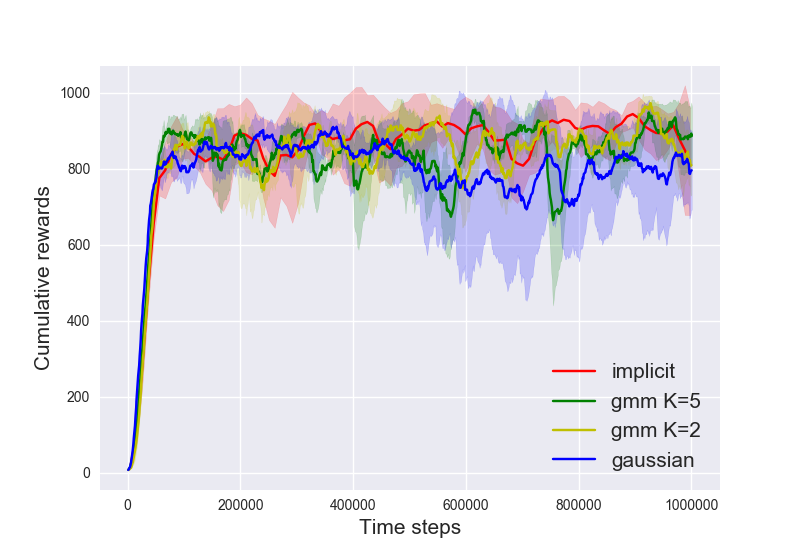}}
\subfigure[\textbf{Double Pendulum}]{\includegraphics[width=.23\linewidth]{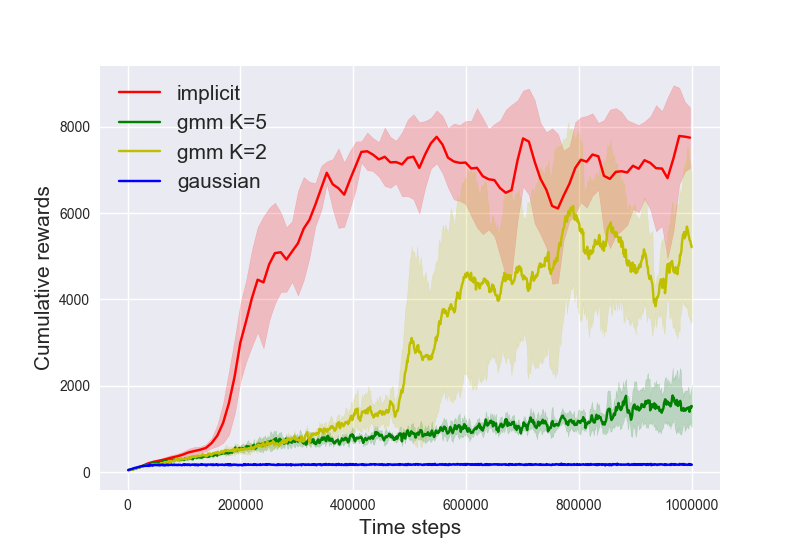}}
\subfigure[\textbf{Hopper}]{\includegraphics[width=.23\linewidth]{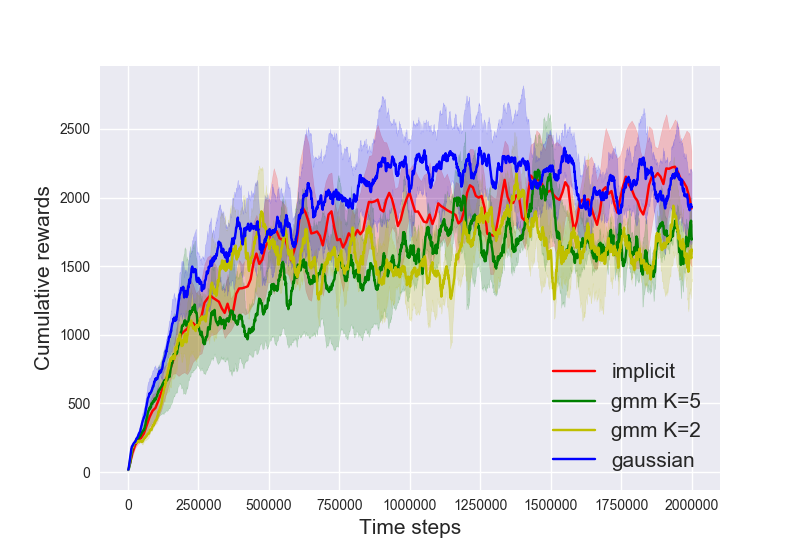}}
\subfigure[\textbf{HalfCheetah}]{\includegraphics[width=.23\linewidth]{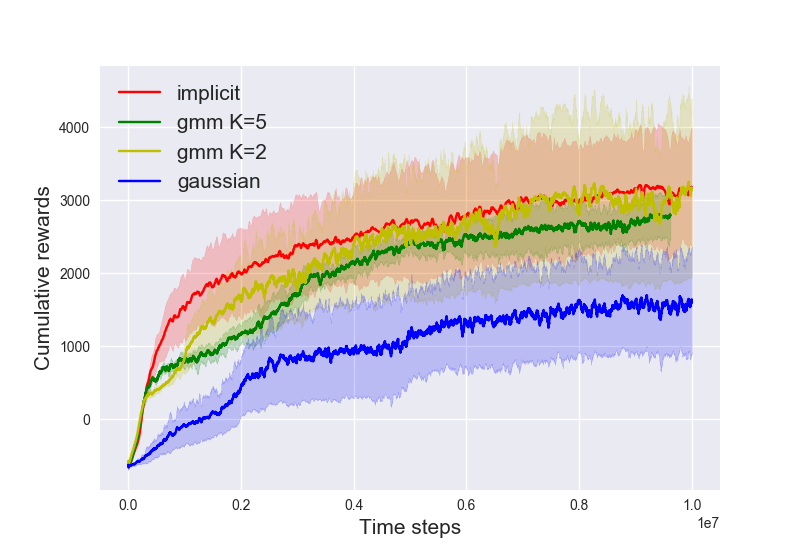}}
\subfigure[\textbf{Ant}]{\includegraphics[width=.23\linewidth]{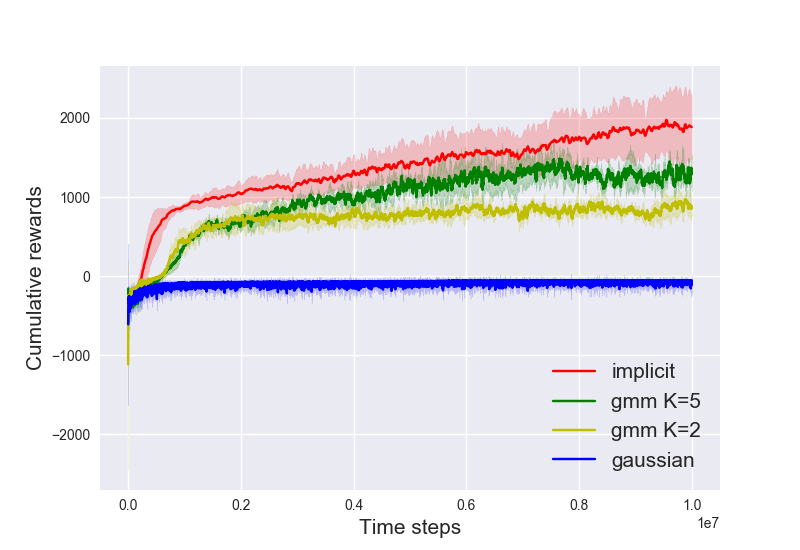}}
\subfigure[\textbf{Walker}]{\includegraphics[width=.23\linewidth]{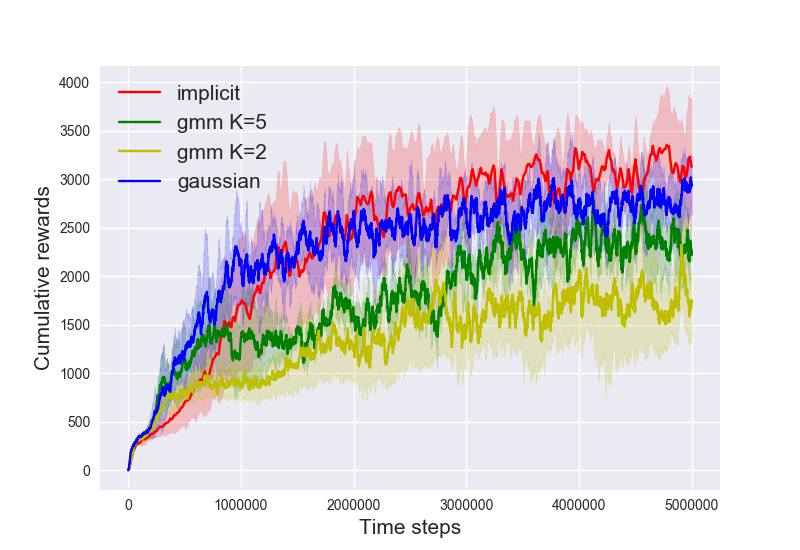}}
\subfigure[\textbf{Sim. Humanoid (L)}]{\includegraphics[width=.23\linewidth]{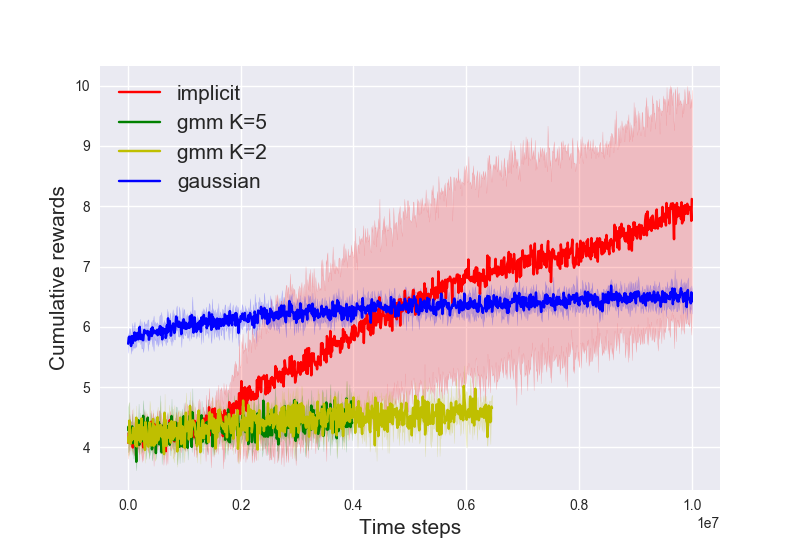}}
\subfigure[\textbf{Humanoid (L)}]{\includegraphics[width=.23\linewidth]{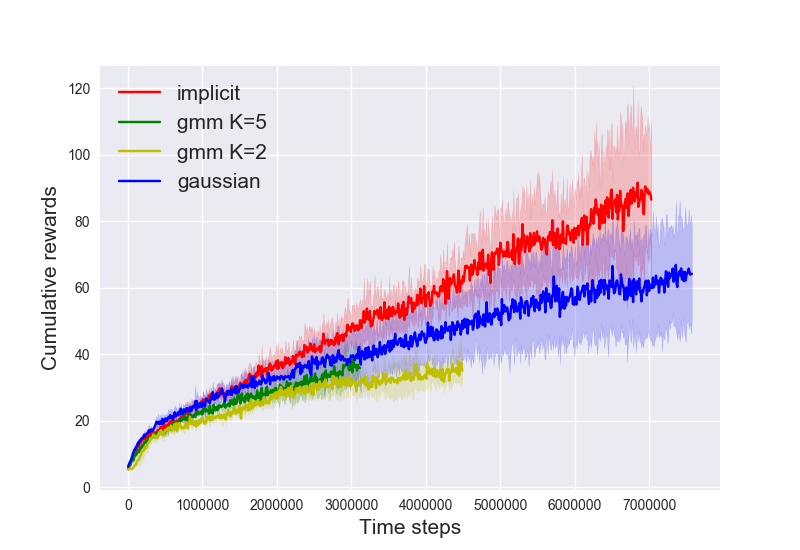}}
\subfigure[\textbf{Humanoid}]{\includegraphics[width=.23\linewidth]{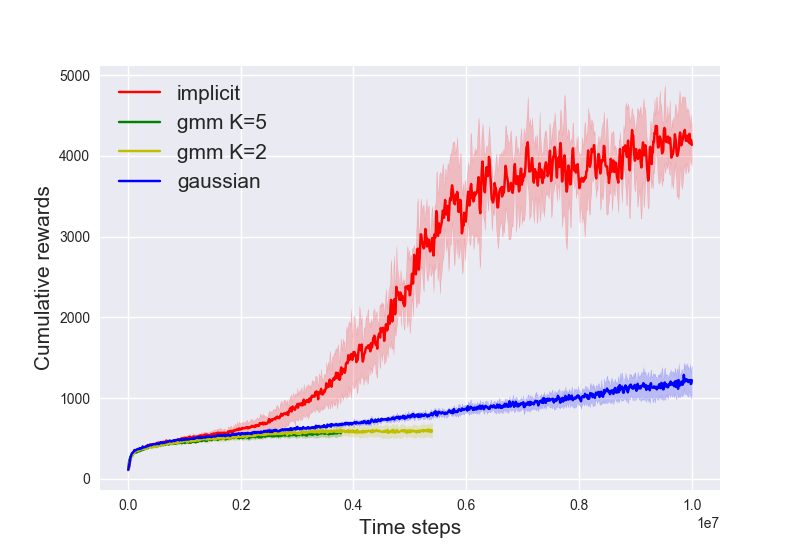}}
\subfigure[\textbf{HumanoidStandup}]{\includegraphics[width=.23\linewidth]{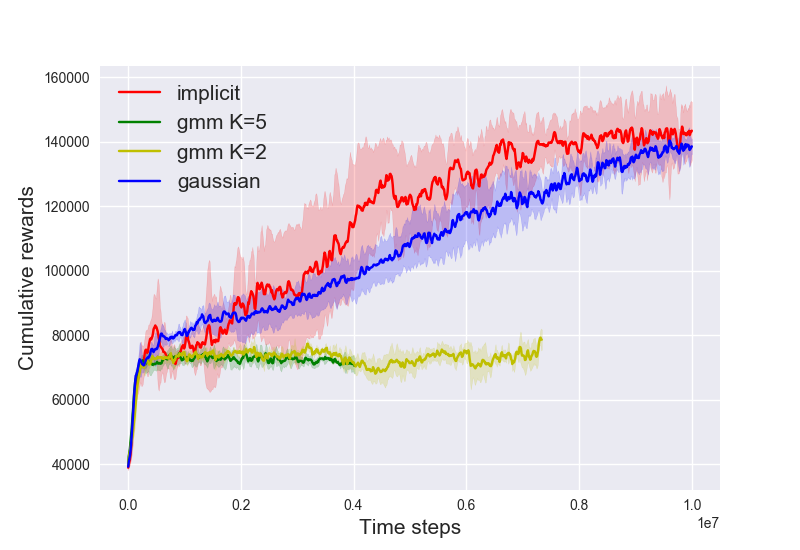}}
\caption{\small{MuJoCo Benchmark: learning curves on MuJoCo locomotion tasks. Tasks with (L) are from rllab. Each curve corresponds to a different policy class (Red: NF (labelled as \emph{implicit}), Green: GMM $K=5$, Yellow: GMM $K=2$, Blue: Gaussian). We observe that the NF policy consistently outperforms other baselines on high-dimensional complex tasks (Bottom two rows).}}
\label{figure:benchmarkdist}
\end{figure*}

\begin{figure}[t]
\centering
\subfigure[\textbf{Humanoid}]{\includegraphics[width=.45\linewidth]{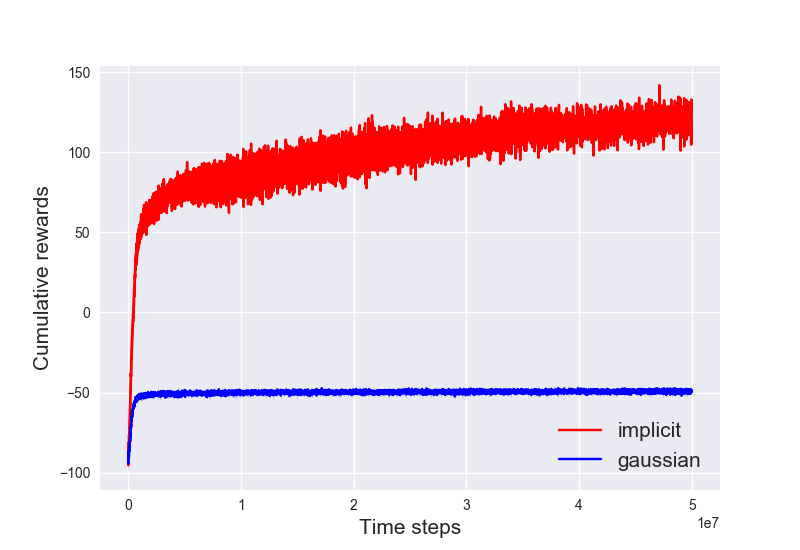}}
\subfigure[\textbf{HumanoidFlagrun}]{\includegraphics[width=.45\linewidth]{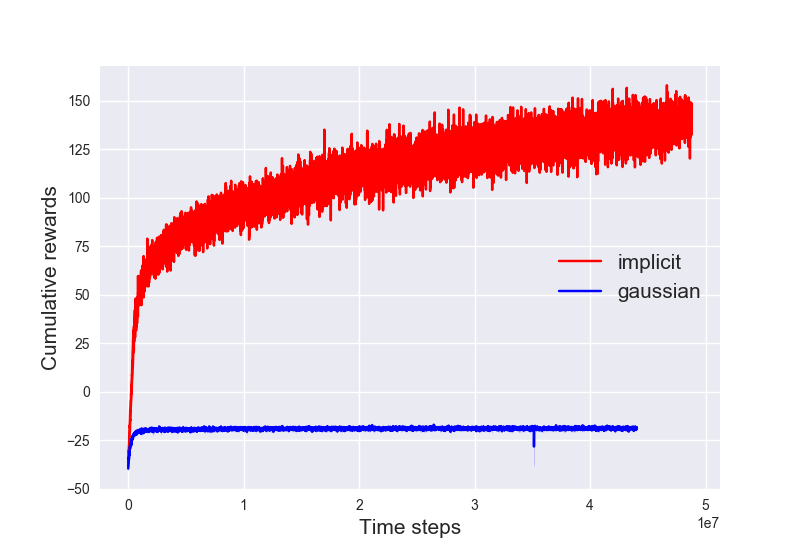}}
\subfigure[\textbf{Flagrun Harder}]{\includegraphics[width=.45\linewidth]{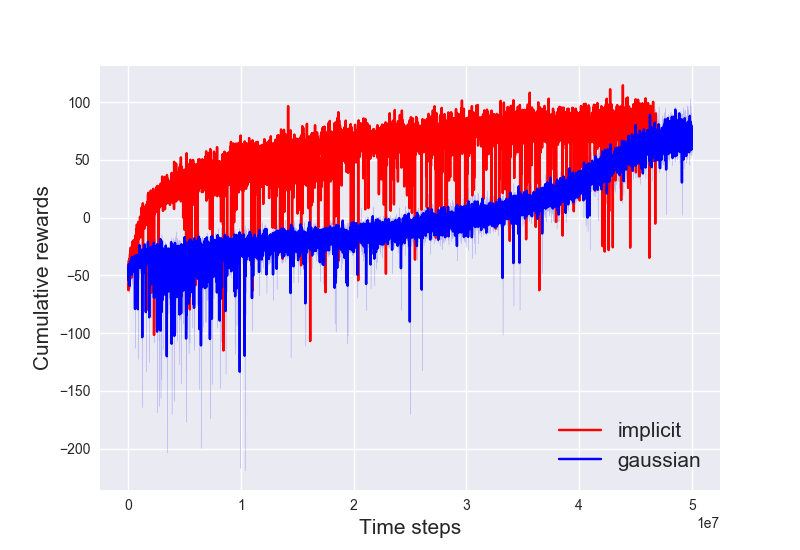}}
\subfigure[\textbf{Illustration}]{\includegraphics[width=.4\linewidth,height=0.9in]{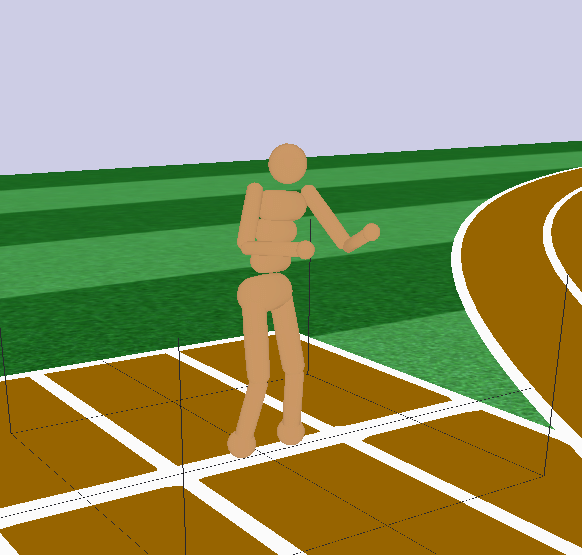}}
\caption{\small{Roboschool Humanoid benchmarks: (a)-(c) show learning curves on Roboschool Humanoid locomotion tasks. Each color corresponds to a different policy class (Red: NF (labelled as \emph{implicit}), Blue: Gaussian). The NF policy significantly outperforms Gaussian since the early stage of training. (d) is an illustration of the Humanoid tasks.}}
\label{figure:roboschool}
\end{figure}

\begin{figure}[t]
\centering
\subfigure[\textbf{BipedalWalker}]{\includegraphics[width=.45\linewidth]{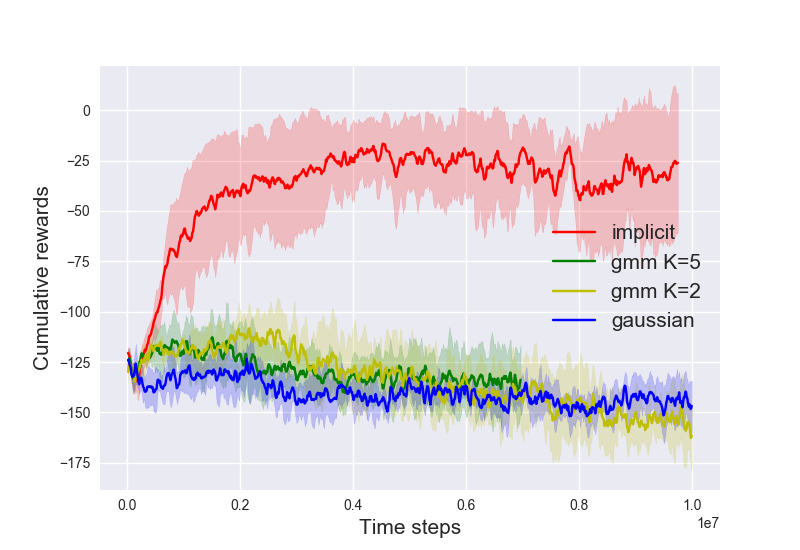}}
\subfigure[\textbf{LunarLander}]{\includegraphics[width=.45\linewidth]{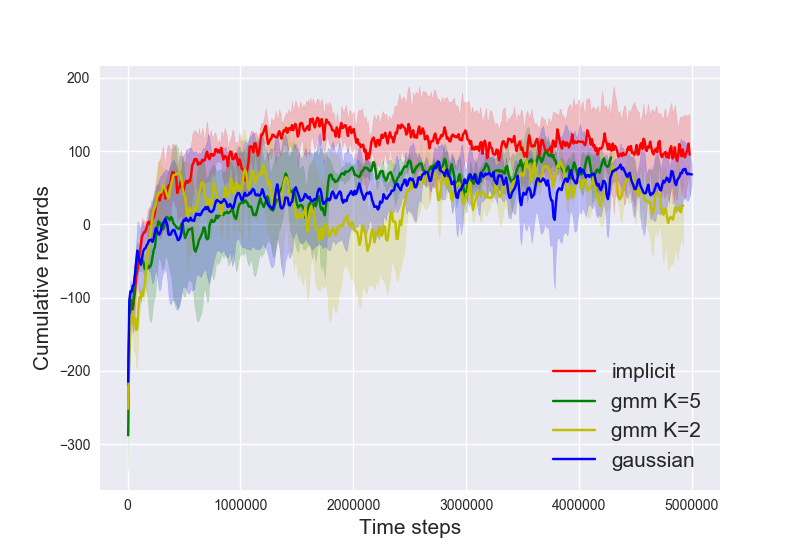}}
\caption{\small{Box2D benchmarks: (a)-(b) show learning curves on Box2D locomotion tasks. Each curve corresponds to a different policy class (Red: NF (labelled as \emph{implicit}), Blue: Gaussian). The NF policy also achieves performance gains on Box2D environments.}}
\label{figure:box2d}
\end{figure}

\begin{figure*}[h]
\centering
\subfigure[\textbf{Walker}]{\includegraphics[width=.23\linewidth]{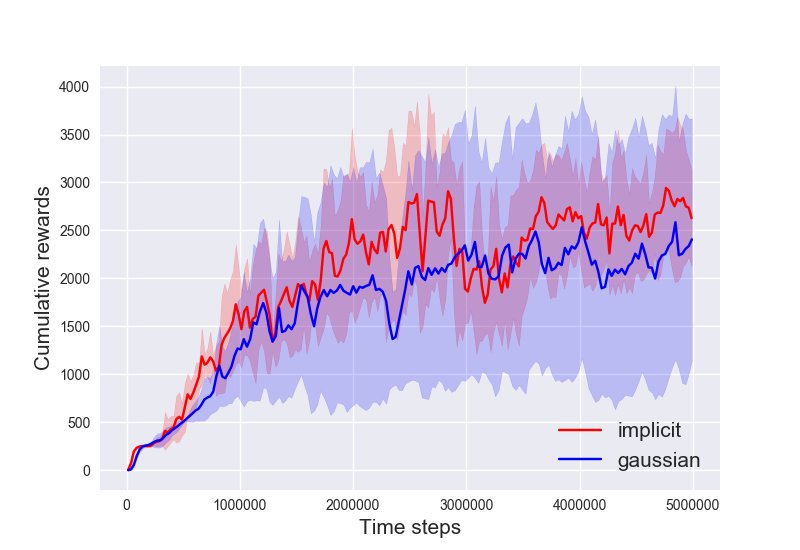}}
\subfigure[\textbf{Walker (R)}]{\includegraphics[width=.23\linewidth]{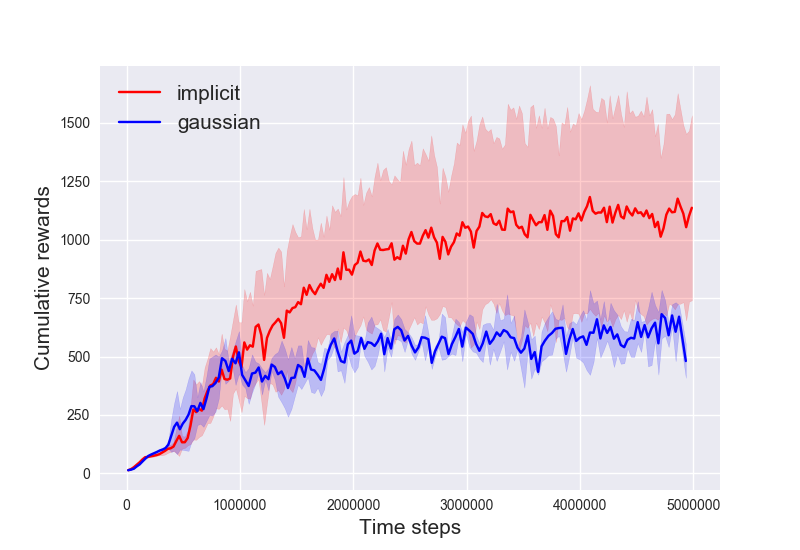}}
\subfigure[\textbf{HalfCheetah}]{\includegraphics[width=.23\linewidth]{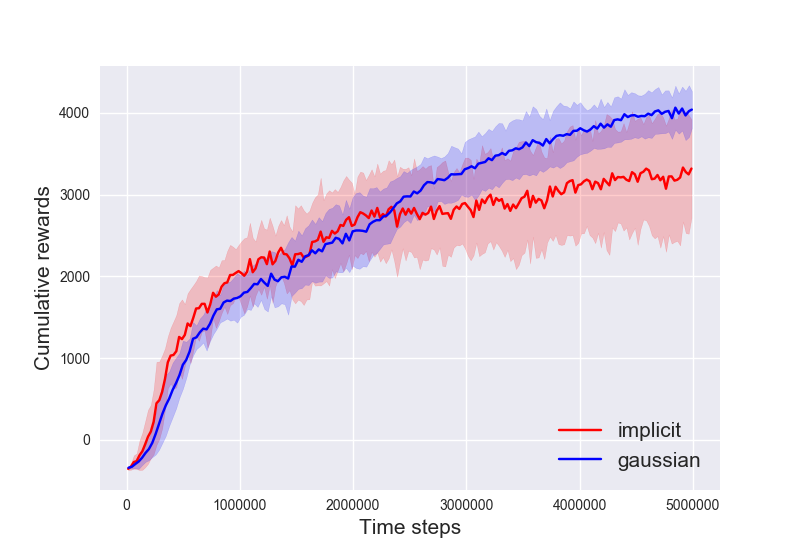}}
\subfigure[\textbf{HalfCheetah (R)}]{\includegraphics[width=.23\linewidth]{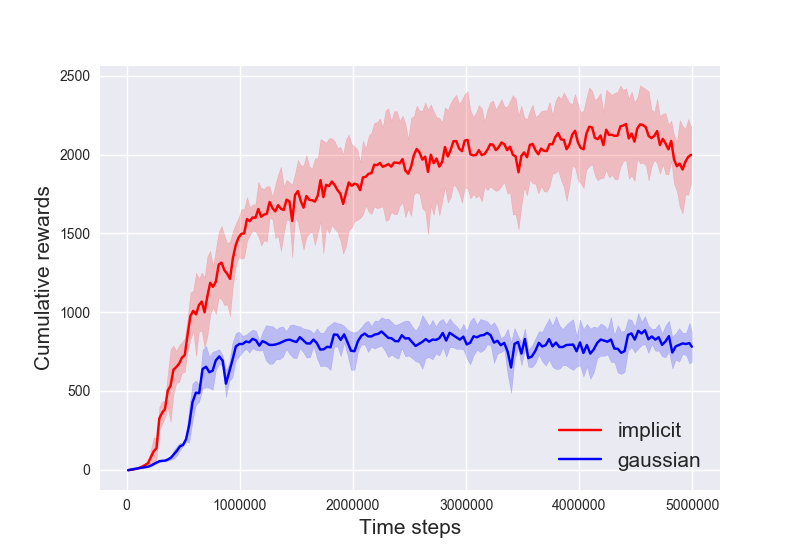}}
\subfigure[\textbf{Ant (R)}]{\includegraphics[width=.23\linewidth]{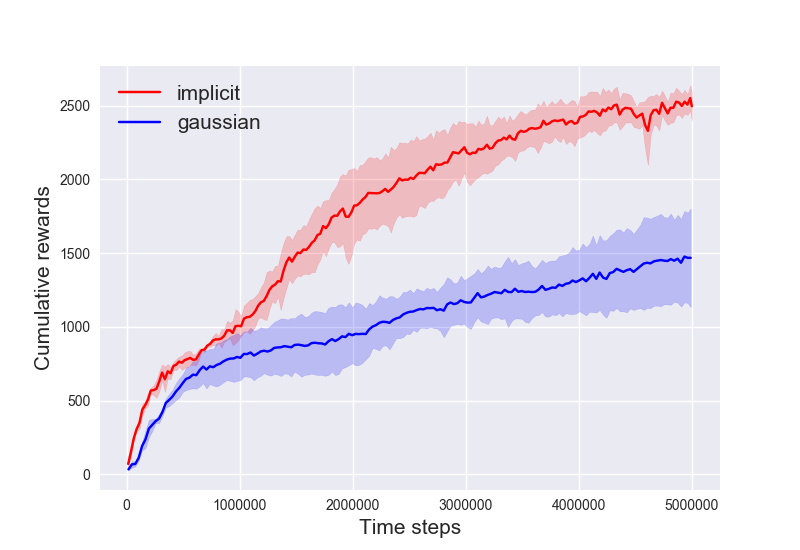}}
\subfigure[\textbf{Sim. Humanoid (L)}]{\includegraphics[width=.23\linewidth]{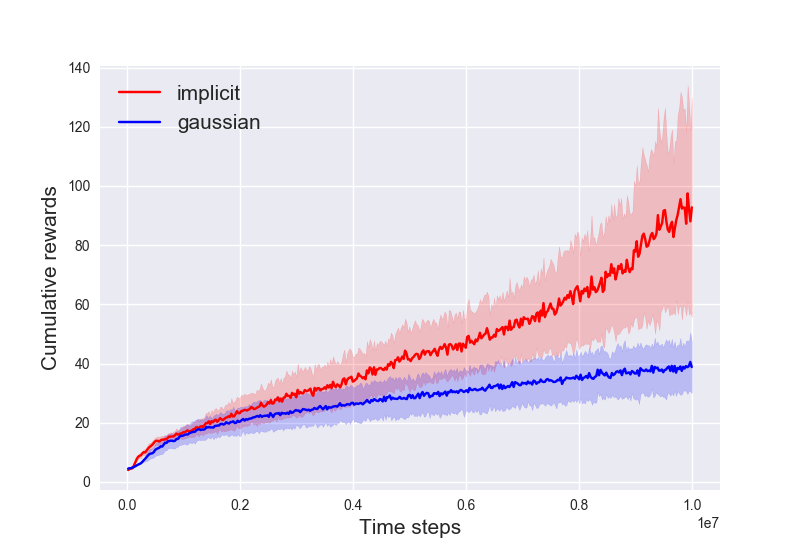}}
\subfigure[\textbf{Humanoid (L)}]{\includegraphics[width=.23\linewidth]{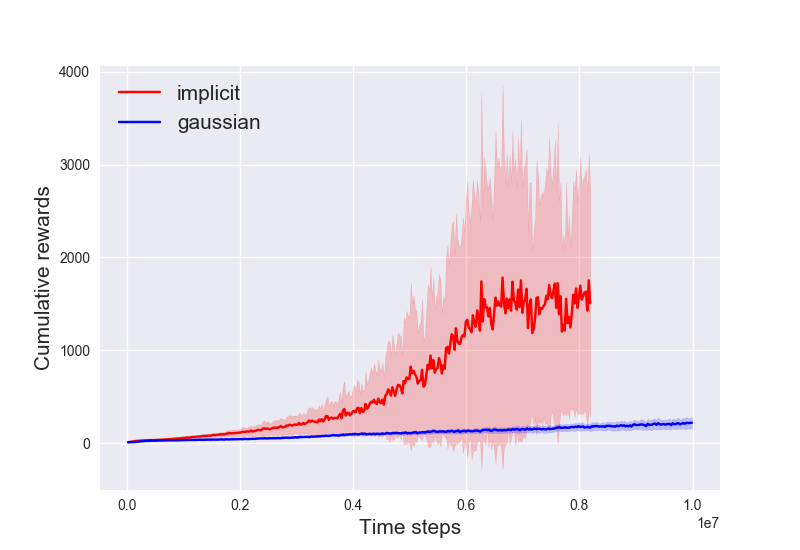}}
\subfigure[\textbf{Humanoid}]{\includegraphics[width=.23\linewidth]{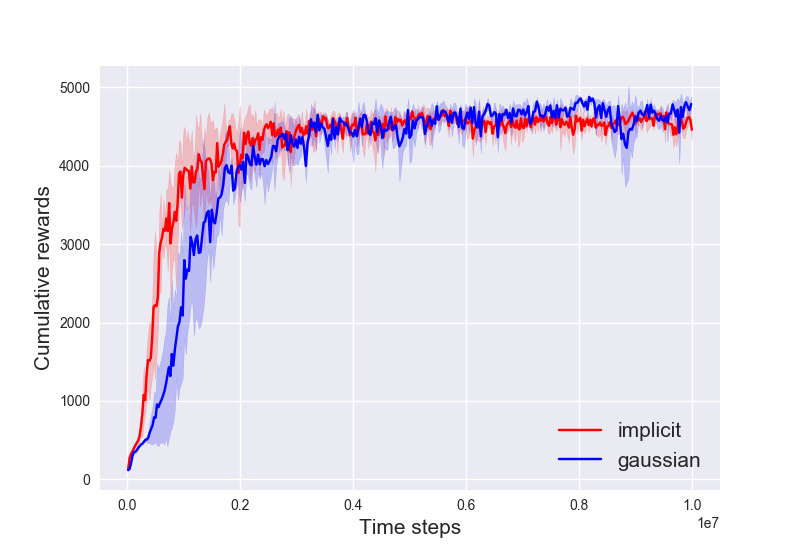}}
\caption{\small{MuJoCo and Roboschool Benchmarks : learning curves on locomotion tasks for ACKTR. Each curve corresponds to a different policy class (Red: NF (labelled as \emph{implicit}), Blue: Gaussian). Tasks with (R) are from Roboschool. The NF policy achieves consistent and significant performance gains across high-dimensional tasks with complex dynamics.}}
\label{figure:benchmarkacktr}
\end{figure*}

All benchmark comparison results are presented in plots (Figure \ref{figure:benchmarkdist},\ref{figure:roboschool},\ref{figure:box2d},\ref{figure:benchmarkacktr}) or tables (Table \ref{table:comparewithotherdists},\ref{table:schulman2017proximal}). For plots, we show the learning curves of different policy classes trained for a fixed number of time steps. The x-axis shows the time steps and the y-axis shows the cumulative rewards. Each curve shows the average performance with standard deviation shown in shaded areas. Results in Figure \ref{figure:benchmarkdist},\ref{figure:box2d},\ref{figure:benchmarkacktr} are averaged over 5 random seeds and Figure \ref{figure:roboschool} over 2 random seeds. In Table \ref{table:comparewithotherdists},\ref{table:schulman2017proximal}, we train all policies for a fixed number of time steps and show the average $\pm$ standard deviation of the cumulative rewards obtained in the last 10 training iterations.

\paragraph{TRPO - Comparison with Gaussian \& GMM.}
In Figure \ref{figure:benchmarkdist} we show the results on benchmark control problems from MuJoCo and in Figure \ref{figure:box2d} we show the results on Box2D tasks. We compare four policy classes under TRPO: factorized Gaussian (blue curves), GMM with $K=2$ (yellow curves), GMM with $K=5$ (green curves) and normalizing flows (red curves). For GMM, each cluster has the same probability weight and each cluster is a factorized Gaussian with independent parameters. We find that though GMM policies with $K\geq 2$ outperform factorized Gaussian on relatively complex tasks such as Ant and HalfCheetah, they suffer from less stable learning in more high-dimensional Humanoid tasks. However, normalizing flows consistently outperforms GMM and factorized Gaussian policies on a wide range of tasks, especially tasks with highly complex dynamics such as Humanoid.

Since we make minimal changes to the algorithm and algorithmic hyper-parameters, we could attribute performance gains to the architectural difference. One critical question is whether the more effective policy optimization results from an increased number of parameters? To study the effect of network size, we train Gaussian policy with large networks (with more parameters than NF policy) and find that this does not lead to performance gains. We detail the results of Gaussian with bigger networks in Appendix C. Such comparison further validates our speculation that the performance gains result from an expressive policy distribution.

\paragraph{TRPO - Comparison with other Architecture Alternatives.}
In Table 1, we compare with some architectural alternatives and recently proposed policy classes: Gaussian distribution with $\text{tanh}$ non-linearity at the output layer, and Beta distribution \citep{chou2017improving}. The primary motivation for these architectures is that they either bound the Gaussian mean (Gaussian $+ \text{tanh}$) or strictly bound the distribution support (Beta distribution), which has been claimed to remove the implicit bias of an unbounded action distribution \citep{chou2017improving}. In Table \ref{table:comparewithotherdists}, We compare the NF policy with these alternatives on benchmark tasks with complex dynamics. We find that the NF policy performs significantly better than other alternatives uniformly and consistently across all presented tasks. 


Here we discuss the results for Beta distribution. In our implementation, we find training with Beta distribution tends to generate numerical errors when the trust region size is large (e.g. $\epsilon = 0.01$). Shrinking the trust region size (e.g. $\epsilon = 0.001$) reduces numerical errors but also greatly degrades the learning performance. We suspect that this is because the Beta distribution parameterization (Appendix A and \citep{chou2017improving}) is  numerically unstable, and we discuss the potential causes in Appendix A. The results in Table \ref{table:comparewithotherdists} for Beta policy are obtained as the performance of the last 10 iterations before termination (potentially prematurely due to numerical error). We make further comparison with Beta policy in Appendix C and show that the NF policy achieves performance gains more consistently and stably.


Our findings suggest that for TRPO, an expressive policy brings more benefits than a policy with a bounded support. We speculate that this is because bounded distributions require warping the sampled actions in a way that makes the optimization more difficult. For example, consider 1-d action space when NF is combined with $\text{tanh}$ nonlinearity for the final output \footnote{Similar to the Gaussian $+\ \text{tanh}$ construct, we can bound the support of the NF policy by applying $\text{tanh}$ at the output, the ending distribution still has tractable likelihood \citep{tuomas2018}.}: the samples $a$ are bounded but an extra factor $(1 - a^2)$ is introduced in the gradient w.r.t. $\theta$. When $a \approx \pm 1$, the gradient can vanish quickly and make it hard to sample on the exact boundary $a = \pm 1$. In practice, we also find the performance of NF policy $+\ \text{tanh}$ to be inferior.

\paragraph{TRPO - Comparison with Gaussian on Roboschool Humanoid and Box2D.}
To further illustrate the strength of the NF policy on high-dimensional tasks, we evaluate normalizing flows vs. factorized Gaussian on Roboschool Humanoid tasks shown in Figure \ref{figure:roboschool}. We observe that ever since the early stage of learning (steps $\leq 10^7$) NF policy (red curves) already outperforms Gaussian (blue curves) by a large margin. In \ref{figure:roboschool} (a)(b), Gaussian is stuck in a locally optimal gait and cannot progress, while the NF policy can bypass such gaits and makes consistent improvement. Surprisingly, BipedalWalker (Hardcore) Box2D  task (Figure \ref{figure:box2d}) is very hard for the Gaussian policy, while the NF policy can make consistent progress during training. 


\begin{table*}[t]
  \caption{\small{Table 1: A comparison of various policy classes on complex benchmark tasks. For each task, we show the cumulative rewards $(\text{mean} \pm \text{std})$ after training for $10^7$ steps across $5$ seeds (for Humanoid (L) it is $7 \cdot 10^6$ steps). For each task, the top two results are highlighted in bold font. The NF policy (labelled as \emph{NF}) consistently achieves top two results.}}
\label{policy}
\vskip 0.15in
\begin{center}
\begin{small}
\begin{sc}
\begin{tabular}{lcccc}
\toprule
 &  Gaussian   &  Gaussian$+ \text{tanh}$ &  Beta &  NF \\ \midrule
 Ant  &  $-76 \pm 14$  & $-89 \pm 13$ & $\mathbf{2362 \pm 305}$ & $\mathbf{1982 \pm 407}$ \\
 HalfCheetah  &  $1576 \pm 782$  & $386 \pm 78$ & $\mathbf{1643 \pm 819}$ & $\mathbf{2900 \pm 554}$ \\
 Humanoid  & $1156 \pm 153$  &  $\mathbf{6350 \pm 486}$  & $3812 \pm 1973$ & $\mathbf{4270 \pm 142}$ \\
 Humanoid (L) & $\mathbf{64.7 \pm 7.6}$  &  $38.2 \pm 2.3$  & $37.8 \pm 3.4$ & $\mathbf{87.2 \pm 19.6}$ \\
 Sim. Humanoid (L) & $\mathbf{6.5 \pm 0.2}$  &  $4.4 \pm 0.1$  & $4.2 \pm 0.2$ & $\mathbf{8.0 \pm 1.8}$ \\
 Humanoid Standup  &  $\mathbf{137955 \pm 9238}$  & $133558 \pm 9238$ & $111497 \pm 15031$ & $\mathbf{142568 \pm 9296}$ \\
\bottomrule
\end{tabular}
\end{sc}
\end{small}
\end{center}
\vskip -0.1in
\label{table:comparewithotherdists}
\end{table*}


\paragraph{TRPO - Comparison with Gaussian results in prior works.} \citet{schulman2017proximal,wu2017scalable} report results on TRPO with Gaussian policy, represented as 2-layer neural network with $64$ hidden units per layer. Since they report the performance after $10^6$ steps of training, we record the performance of NF policy after $10^6$ steps of training for fair comparison in Table \ref{table:schulman2017proximal}. The results of \citep{schulman2017proximal,wu2017scalable} in Table \ref{table:schulman2017proximal} are directly estimated from the plots in their paper, following the practice of \citep{mania2018simple}. We see that our proposed method outperforms the Gaussian policy $+$ TRPO reported in \citep{schulman2017proximal,wu2017scalable} for most control tasks.

Some prior works \citep{tuomas2018} also report the results for Gaussian policy after training for more steps. We also offer such a comparison in Appendix C, where we show that the NF policy still achieves significantly better results on reported tasks.

Comprehensive comparison across results reported in prior works ensures that we compare with (approximate) state-of-the-art results. Despite implementation and hyper-parameter differences from various prior works, the NF policy can consistently achieve faster rate of learning and better asymptotic performance.


\paragraph{ACKTR - Comparison with Gaussian.}
We also evaluate different policy classes combined with ACKTR \citep{wu2017scalable}. In Figure \ref{figure:benchmarkacktr}, we compare factorized Gaussian (red curves) against NF (blue curves) on a suite of MuJoCo and Roboschool control tasks. Though the NF policy does not uniformly outperform Gaussian on all tasks, we find that for tasks with relatively complex dynamics (e.g. Ant and Humanoid), the NF policy achieves significant performance gains. We find that the effect of an expressive policy class is fairly orthogonal to the additional training stability introduced by ACKTR over TRPO and the combined algorithm achieves even better performance.


\begin{table*}[t]
\caption{Comparison with TRPO results reported in \citep{schulman2017proximal,wu2017scalable}. Since \citet{schulman2017proximal,wu2017scalable} report the performance for $10^6$ steps of training, we also record the performance of our method after training for $10^6$ for fair comparison. Better results are highlighted in bold font. When results are not statistically different, both are highlighted.}
\label{policy}
\vskip 0.15in
\begin{center}
\begin{small}
\begin{sc}
\begin{tabular}{lcccc}
\toprule
Task & NF (ours) & Gaussian (ours) & Schulman et al. 2017 & Wu et al. 2017\\
\midrule
Reacher  & $\mathbf{\approx -10}$ & $\approx -115$ & $\approx -115$ & $\mathbf{\approx -10}$  \\
Swimmer & $\approx 64$ & $\approx 90$ &  $\mathbf{\approx 120}$ & $\approx 40$ \\
Inverted Pend. & $\approx 900$ &   $\approx 800$ & $\approx 900$ & $\mathbf{\approx 1000}$ \\
Double Pend. & $\mathbf{\approx 7800}$ & $<1000$ &  $\approx 0$ & $< 1000$ \\
Hopper & $\mathbf{\approx 2000}$ & $\mathbf{\approx 2000}$ & $\mathbf{\approx 2000}$ & $\approx 1400$\\
HalfCheetah & $\mathbf{\approx 1500}$ & $\approx 0$ & $\approx 0$ & $< 500$  \\
Walker2d & $\approx 1700$ & $\mathbf{\approx 2000}$ & $\approx 1000$ & $\approx 550$   \\
Ant & $\mathbf{\approx 800}$ & $\approx 0$ & N/A & $<-500$ \\
\bottomrule
\end{tabular}
\end{sc}
\end{small}
\end{center}
\vskip -0.1in
\label{table:schulman2017proximal}
\end{table*}



\subsection{Sensitivity to Hyper-parameters and Ablation Study}
We evaluate the policy classes' sensitivities to hyper-parameters in Appendix C, where we compare Gaussian vs. normalizing flows. Recall that $\epsilon$ is the constant for KL constraint. For each policy, we uniformly random sample $\log_{10}\epsilon \in [-3.0,-2.0]$ and one of five random seeds, and train policies with TRPO for a fixed number of time steps. The final performance (cumulative rewards) is recorded and Figure \ref{figure:hyperparams} in Appendix C shows the quantile plots of final rewards across multiple tasks. We observe that the NF policy is generally much more robust to such hyper-parameters, importantly to $\epsilon$. We speculate that such additional robustness partially stems from the fact that for the NF policy, the KL constraint does not pose very stringent restriction on the sampled action space, which allows the system to efficiently explore even when $\epsilon$ is small.

We carry out a small ablation study that addresses how hyper-parameters inherent to normalizing flows can impact the results. Recall that the NF policy (Section 4) consists of $K$ transformations, with the first transformation embedding the state $s$ into a vector $L_{\theta_s}(s)$. Here we implement $L_{\theta_s}(s)$ as a two-layer neural networks with $l_1$ hidden units per layer. We evaluate on the policy performance as we vary $K \in \{2,4,6\}$ and $l_1 \in \{3,5,7\}$. We find that the performance of NF policies are fairly robust to such hyper-parameters (see Appendix C).

\section{Conclusion}
We propose normalizing flows as a novel on-policy architecture to boost the performance of trust region policy search. In particular, we observe that the empirical properties of the NF policy allows for better exploration in practice. We evaluate performance of NF policy combined with trust region algorithms (TRPO/ ACKTR) and show that they consistently and significantly outperform a large number of previous policy architectures. We have not observed similar performance gains on gradient based methods such as PPO \citep{schulman2017} and we leave this as future work.

\section{Acknowledgements}
This work was supported by an Amazon Research Award (2017).  The authors also acknowledge the computational resources provided by Amazon Web Services (AWS).




\bibliography{iclr2019_conference}

\begin{thebibliography}{27}
\providecommand{\natexlab}[1]{#1}
\providecommand{\url}[1]{\texttt{#1}}
\expandafter\ifx\csname urlstyle\endcsname\relax
  \providecommand{\doi}[1]{doi: #1}\else
  \providecommand{\doi}{doi: \begingroup \urlstyle{rm}\Url}\fi

\bibitem[Brockman et~al.(2016)Brockman, Cheung, Pettersson, Schneider,
  Schulman, Tang, and Zaremba]{brockman2016}
Greg Brockman, Vicki Cheung, Ludwig Pettersson, Jonas Schneider, John Schulman,
  Jie Tang, and Wojciech Zaremba.
\newblock Openai gym.
\newblock \emph{arXiv preprint arXiv:1606.01540}, 2016.

\bibitem[Chou et~al.(2017)Chou, Maturana, and Scherer]{chou2017improving}
Po-Wei Chou, Daniel Maturana, and Sebastian Scherer.
\newblock Improving stochastic policy gradients in continuous control with deep
  reinforcement learning using the beta distribution.
\newblock In \emph{International Conference on Machine Learning}, pp.\
  834--843, 2017.

\bibitem[Dhariwal et~al.(2017)Dhariwal, Hesse, Klimov, Nichol, Plappert,
  Radford, Schulman, Sidor, and Wu]{baselines}
Prafulla Dhariwal, Christopher Hesse, Oleg Klimov, Alex Nichol, Matthias
  Plappert, Alec Radford, John Schulman, Szymon Sidor, and Yuhuai Wu.
\newblock Openai baselines.
\newblock \url{https://github.com/openai/baselines}, 2017.

\bibitem[Dinh et~al.(2014)Dinh, Krueger, and Bengio]{dinh2015}
Laurent Dinh, David Krueger, and Yoshua Bengio.
\newblock Nice: Non-linear independent components estimation.
\newblock \emph{arXiv preprint arXiv:1410.8516}, 2014.

\bibitem[Dinh et~al.(2016)Dinh, Sohl-Dickstein, and Bengio]{dinh2017}
Laurent Dinh, Jascha Sohl-Dickstein, and Samy Bengio.
\newblock Density estimation using real nvp.
\newblock \emph{arXiv preprint arXiv:1605.08803}, 2016.

\bibitem[Duan et~al.(2016)Duan, Chen, Houthooft, Schulman, and
  Abbeel]{duanxi2016}
Yan Duan, Xi~Chen, Rein Houthooft, John Schulman, and Pieter Abbeel.
\newblock Benchmarking deep reinforcement learning for continuous control.
\newblock In \emph{International Conference on Machine Learning}, pp.\
  1329--1338, 2016.

\bibitem[Haarnoja et~al.(2017)Haarnoja, Tang, Abbeel, and Levine]{tuomas2017}
Tuomas Haarnoja, Haoran Tang, Pieter Abbeel, and Sergey Levine.
\newblock Reinforcement learning with deep energy-based policies.
\newblock \emph{arXiv preprint arXiv:1702.08165}, 2017.

\bibitem[Haarnoja et~al.(2018{\natexlab{a}})Haarnoja, Hartikainen, Abbeel, and
  Levine]{tuomas2018b}
Tuomas Haarnoja, Kristian Hartikainen, Pieter Abbeel, and Sergey Levine.
\newblock Latent space policies for hierarchical reinforcement learning.
\newblock \emph{arXiv preprint arXiv:1804.02808}, 2018{\natexlab{a}}.

\bibitem[Haarnoja et~al.(2018{\natexlab{b}})Haarnoja, Zhou, Abbeel, and
  Levine]{tuomas2018}
Tuomas Haarnoja, Aurick Zhou, Pieter Abbeel, and Sergey Levine.
\newblock Soft actor-critic: Off-policy maximum entropy deep reinforcement
  learning with a stochastic actor.
\newblock \emph{arXiv preprint arXiv:1801.01290}, 2018{\natexlab{b}}.

\bibitem[Henderson et~al.(2017)Henderson, Islam, Bachman, Pineau, Precup, and
  Meger]{henderson2017deep}
Peter Henderson, Riashat Islam, Philip Bachman, Joelle Pineau, Doina Precup,
  and David Meger.
\newblock Deep reinforcement learning that matters.
\newblock \emph{arXiv preprint arXiv:1709.06560}, 2017.

\bibitem[Kakade \& Langford(2002)Kakade and Langford]{kakade2002approximately}
Sham Kakade and John Langford.
\newblock Approximately optimal approximate reinforcement learning.
\newblock In \emph{ICML}, volume~2, pp.\  267--274, 2002.

\bibitem[Kakade(2002)]{kakade2002natural}
Sham~M Kakade.
\newblock A natural policy gradient.
\newblock In \emph{Advances in neural information processing systems}, pp.\
  1531--1538, 2002.

\bibitem[Kingma \& Dhariwal(2018)Kingma and Dhariwal]{kingma2018glow}
Diederik~P Kingma and Prafulla Dhariwal.
\newblock Glow: Generative flow with invertible 1x1 convolutions.
\newblock \emph{arXiv preprint arXiv:1807.03039}, 2018.

\bibitem[Levine(2018)]{levine2018reinforcement}
Sergey Levine.
\newblock Reinforcement learning and control as probabilistic inference:
  Tutorial and review.
\newblock \emph{arXiv preprint arXiv:1805.00909}, 2018.

\bibitem[Liu \& Wang(2016)Liu and Wang]{liu2016}
Qiang Liu and Dilin Wang.
\newblock Stein variational gradient descent: A general purpose bayesian
  inference algorithm.
\newblock In \emph{Advances In Neural Information Processing Systems}, pp.\
  2378--2386, 2016.

\bibitem[Louizos \& Welling(2017)Louizos and
  Welling]{louizos2017multiplicative}
Christos Louizos and Max Welling.
\newblock Multiplicative normalizing flows for variational bayesian neural
  networks.
\newblock \emph{arXiv preprint arXiv:1703.01961}, 2017.

\bibitem[Mania et~al.(2018)Mania, Guy, and Recht]{mania2018simple}
Horia Mania, Aurelia Guy, and Benjamin Recht.
\newblock Simple random search provides a competitive approach to reinforcement
  learning.
\newblock \emph{arXiv preprint arXiv:1803.07055}, 2018.

\bibitem[Martens \& Grosse(2015)Martens and Grosse]{martens2015optimizing}
James Martens and Roger Grosse.
\newblock Optimizing neural networks with kronecker-factored approximate
  curvature.
\newblock In \emph{International conference on machine learning}, pp.\
  2408--2417, 2015.

\bibitem[Mnih et~al.(2016)Mnih, Badia, Mirza, Graves, Lillicrap, Harley,
  Silver, and Kavukcuoglu]{mnih2016}
Volodymyr Mnih, Adria~Puigdomenech Badia, Mehdi Mirza, Alex Graves, Timothy
  Lillicrap, Tim Harley, David Silver, and Koray Kavukcuoglu.
\newblock Asynchronous methods for deep reinforcement learning.
\newblock In \emph{International Conference on Machine Learning}, pp.\
  1928--1937, 2016.

\bibitem[Rezende \& Mohamed(2015)Rezende and Mohamed]{rezende2015}
Danilo~Jimenez Rezende and Shakir Mohamed.
\newblock Variational inference with normalizing flows.
\newblock \emph{arXiv preprint arXiv:1505.05770}, 2015.

\bibitem[Schulman et~al.(2015)Schulman, Levine, Abbeel, Jordan, and
  Moritz]{schulman2015}
John Schulman, Sergey Levine, Pieter Abbeel, Michael Jordan, and Philipp
  Moritz.
\newblock Trust region policy optimization.
\newblock In \emph{International Conference on Machine Learning}, pp.\
  1889--1897, 2015.

\bibitem[Schulman et~al.(2017{\natexlab{a}})Schulman, Wolski, Dhariwal,
  Radford, and Klimov]{schulman2017}
John Schulman, Filip Wolski, Prafulla Dhariwal, Alec Radford, and Oleg Klimov.
\newblock Proximal policy optimization algorithms.
\newblock \emph{arXiv preprint arXiv:1707.06347}, 2017{\natexlab{a}}.

\bibitem[Schulman et~al.(2017{\natexlab{b}})Schulman, Wolski, Dhariwal,
  Radford, and Klimov]{schulman2017proximal}
John Schulman, Filip Wolski, Prafulla Dhariwal, Alec Radford, and Oleg Klimov.
\newblock Proximal policy optimization algorithms.
\newblock \emph{arXiv preprint arXiv:1707.06347}, 2017{\natexlab{b}}.

\bibitem[Tang \& Agrawal(2018)Tang and Agrawal]{tang2018implicit}
Yunhao Tang and Shipra Agrawal.
\newblock Implicit policy for reinforcement learning.
\newblock \emph{arXiv preprint arXiv:1806.06798}, 2018.

\bibitem[Todorov(2008)]{todorov2008}
Emanuel Todorov.
\newblock General duality between optimal control and estimation.
\newblock In \emph{Decision and Control, 2008. CDC 2008. 47th IEEE Conference
  on}, pp.\  4286--4292. IEEE, 2008.

\bibitem[Wright \& Nocedal(1999)Wright and Nocedal]{wright1999numerical}
Stephen Wright and Jorge Nocedal.
\newblock Numerical optimization.
\newblock \emph{Springer Science}, 35\penalty0 (67-68):\penalty0 7, 1999.

\bibitem[Wu et~al.()Wu, Mansimov, Gross, Liao, and Ba]{wu2017scalable}
Yuhuai Wu, Elman Mansimov, Roger~B Gross, Shun Liao, and Jummy Ba.
\newblock Scalable trust-region method for deep reinforcement learning using
  kronecker-factored approximation.
\newblock \emph{Advances in neural information processing systems}, pp.\
  5279--5288.

\end{thebibliography}
\bibliographystyle{iclr2019_conference}

\newpage
\onecolumn

\appendix
\section{Hyper-parameters}
 All implementations of algorithms (TRPO and ACKTR) are based on OpenAI baselines \citep{baselines}. We implement our own GMM policy and NF policy. Environments are based on OpenAI gym \citep{brockman2016}, rllab \citep{duanxi2016} and Roboschool \citep{schulman2017proximal}. 

 We remark that various policy classes have exactly the same interface to TRPO and ACKTR. In particular, TRPO and ACKTR only requires the computation of $\log \pi_\theta(a|s)$ (and its derivative). Different policy classes only differ in how they parameterize $\pi_\theta(a|s)$ and can be easily plugged into the algorithmic procedure originally designed for Gaussian \citep{baselines}.

We present the details of each policy class as follows.

\paragraph{Factorized Gaussian Policy.} A factorized Gaussian policy has the form $\pi_\theta(\cdot|s) = \mathbb{N}(\mu_\theta(s), \Sigma)$, where $\Sigma$ is a diagonal matrix with $\Sigma_{ii} = \sigma_i^2$. We use the default hyper-parameters in baselines for factorized Gaussian policy. The mean $\mu_\theta(s)$ parameterized by a two-layer neural network with $64$ hidden units per layer and $\text{tanh}$ activation function. The standard deviation $\sigma_i^2$ is each a single variable shared across all states.

\paragraph{Factorized Gaussian$+ \text{tanh}$ Policy.} The architecture is the same as above but the final layer is added a $\tanh$ transformation to ensure that the mean $\mu_\theta(s) \in [-1,1]$.

\paragraph{GMM Policy.} A GMM policy has the form $\pi_\theta(\cdot|s) = \sum_{i=1}^K p_i \mathbb{N}(\mu^{(i)}_\theta(s),\Sigma_i)$, where the cluster weight $p_i = \frac{1}{K}$ is fixed and $\mu_\theta^{(i)}(s),\Sigma_i$ are Gaussian parameters for the $i$th cluster. Each Gaussian has the same parameterization as the factorized Gaussian above.

\paragraph{Beta Policy.} A Beta policy has the form $\pi(\alpha_\theta(s),\beta_\theta(s))$ where $\pi$ is a Beta distribution with parameters $\alpha_\theta(s),\beta_\theta(s)$. Here, $\alpha_\theta(s)$ and $\beta_\theta(s)$ are shape/rate parameters parameterized by two-layer neural network $f_\theta(s)$ with a softplus at the end, i.e. $\alpha_\theta(s) = \log(\exp(f_\theta(s)) + 1) + 1$, following \citep{chou2017improving}. Actions sampled from this distribution have a strictly finite support. We notice that this parameterization introduces potential instability during optimization: for example, when we want to converge on policies that sample actions at the boundary, we require $\alpha_\theta(s) \rightarrow \infty$ or $\beta_\theta(s) \rightarrow \infty$, which might be very unstable. We also observe such instability in practice: when the trust region size is large (e.g. $\epsilon = 0.01$) the training can easily terminate prematurely due to numerical errors. However, reducing the trust region size (e.g. $\epsilon = 0.001$) will stabilize the training but degrade the performance.

\paragraph{Normalizing flows Policy (NF Policy).} A NF policy has a generative form: the sample $a \sim \pi_\theta(\cdot|s)$ can be generated via $a = f_\theta(s,\epsilon)$ with $\epsilon \sim \rho_0(\cdot)$. The detailed architecture of $f_\theta(s,\epsilon)$ is in Appendix B.

\paragraph{Other Hyper-parameters.} Value functions are all parameterized as two-layer neural networks with 64 hidden units per layer and $\text{tanh}$ activation function. Trust region sizes are enforced via a constraint parameter $\epsilon$, where $\epsilon \in \{0.01,0.001\}$ for TRPO and $\epsilon \in \{0.02,0.002\}$ for ACKTR. All other hyper-parameters are default parameters from the baselines implementations.

\section{Normalizing flows Policy Architecture}
We design the neural network architecture following the idea of \citep{dinh2015,dinh2017}. Recall that normalizing flows \citep{rezende2015} consists of layers of transformation as follows
,$$
x = g_{\theta_K} \circ g_{\theta_{K-1}} \circ ... \circ g_{\theta_2} \circ g_{\theta_1} (\epsilon),
$$
where each $g_{\theta_i}(\cdot)$ is an invertible transformation. We focus on how to design each atomic transformation $g_{\theta_i}(\cdot)$. We overload the notations and let $x,y$ be the input/output of a generic layer $g_{\theta}(\cdot)$, 
$$y = g_{\theta}(x).$$
We design a generic transformation $g_\theta(\cdot)$ as follows. Let $x_{I}$ be the components of $x$ corresponding to subset indices $I \subset \{1,2...m\}$. Then we propose as in \citep{dinh2017},
\begin{align}
y_{1:d} &= x_{1:d} \nonumber \\ 
y_{d+1:m} &= x_{d+1:m} \odot \exp(s(x_{1:d})) + t(x_{1:d}),
\label{eq:couplinglayer}
\end{align}
where $t(\cdot),s(\cdot)$ are two arbitrary functions $t,s : \mathbb{R}^{d} \mapsto \mathbb{R}^{m-d}$. It can be shown that such transformation entails a simple Jacobien matrix
$|\frac{\partial y}{\partial x^T}| = \exp(\sum_{j=1}^{m-d} [s(x_{1:d})]_j)$ where $[s(x_{1:d})]_j$ refers to the $j$th component of $s(x_{1:d})$ for $1\leq j\leq m-d$. For each layer, we can permute the input $x$ before apply the simple transformation (\ref{eq:couplinglayer}) so as to couple different components across layers. Such coupling entails a complex transformation when we stack multiple layers of (\ref{eq:couplinglayer}). To define a policy, we need to incorporate state information. We propose to preprocess the state $s \in \mathbb{R}^n$ by a neural network $L_{\theta_s}(\cdot)$ with parameter $\theta_s$, to get a state vector $L_{\theta_s}(s) \in \mathbb{R}^m$. Then combine the state vector into (\ref{eq:couplinglayer}) as follows,
\begin{align}
z_{1:d} &= x_{1:d} \nonumber \\ 
z_{d+1:m} &= x_{d+1:m} \odot \exp(s(x_{1:d})) + t(x_{1:d}) \nonumber \\
y &= z + L_{\theta_s}(s).
\label{eq:statecouplinglayer}
\end{align}

It is obvious that $x \leftrightarrow y$ is still bijective regardless of the form of $L_{\theta_s}(\cdot)$ and the Jacobien matrix is easy to compute accordingly. 

In locomotion benchmark experiments, we implement $s,t$ both as 4-layers neural networks with $l_1=3$ units per hidden layer. We stack $K=4$ transformations: we implement (\ref{eq:statecouplinglayer}) to inject state information only after the first transformation, and the rest is conventional coupling as in (\ref{eq:couplinglayer}). $L_{\theta_s}(s)$ is implemented as a feedforward neural network with $2$ hidden layers each with $64$ hidden units. Value function critic is implemented as a feedforward neural network with $2$ hidden layers each with $64$ hidden units with rectified-linear between hidden layers.

\section{Additional Experiments}
\subsection{Sensitivity to Hyper-Parameters and Ablation Study}
In Figure \ref{figure:ablation}, we show the ablation study of the NF policy. We evaluate how the training curves change as we change the hyper-parameters of NF policy: number of transformation $K\in\{2,4,6\}$ and number of hidden units $l_1\in\{3,5,7\}$ in the embedding function $L_{\theta_s}(\cdot)$. We find that the performance of the NF policy is fairly robust to changes in $K$ and $l_1$. When $K$ varies, $l_1$ is set to $3$ by default. When $l_1$ varies, $K$ is set to $4$ by default.
 
\begin{figure}[h]
\centering
\subfigure[\textbf{Reacher}]{\includegraphics[width=.23\linewidth]{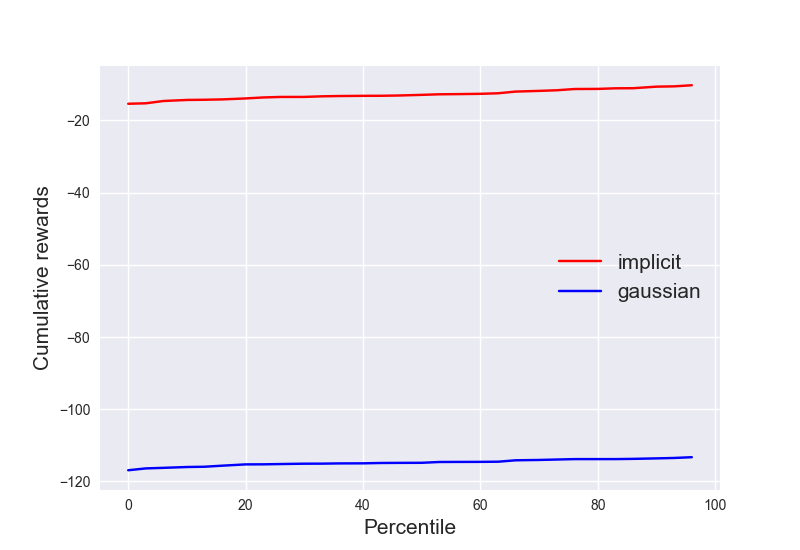}}
\subfigure[\textbf{Hopper}]{\includegraphics[width=.23\linewidth]{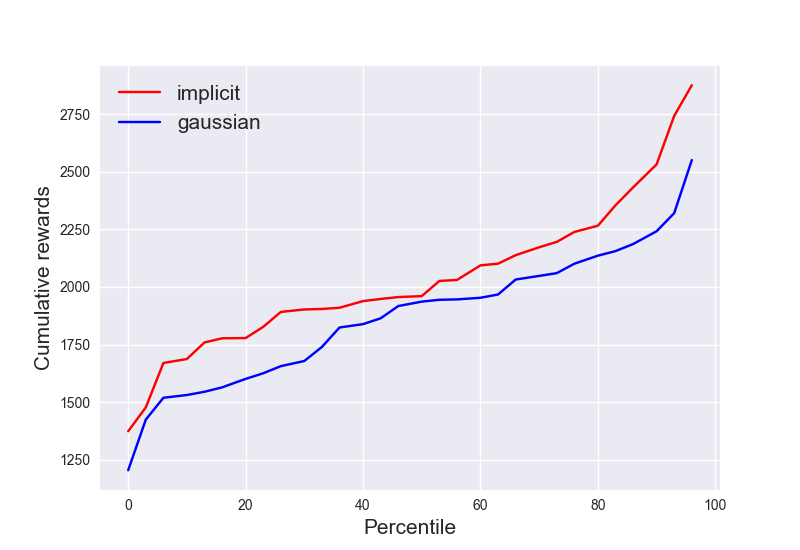}}
\subfigure[\textbf{HalfCheetah}]{\includegraphics[width=.23\linewidth]{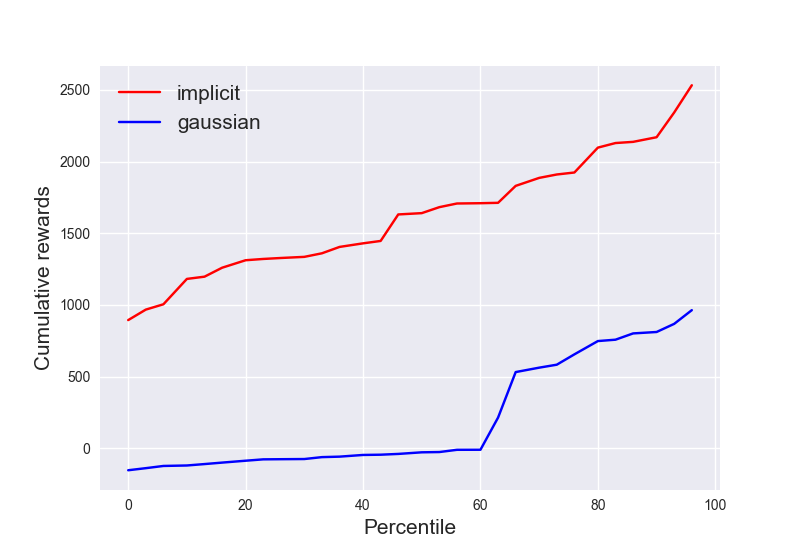}}
\subfigure[\textbf{Sim. Humanoid}]{\includegraphics[width=.23\linewidth]{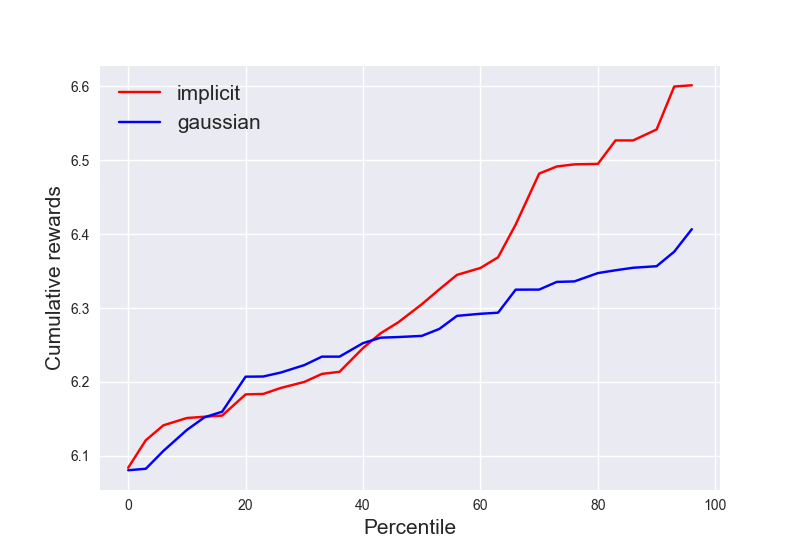}}
\caption{\small{Sensitivity to Hyper-parameters: quantile plots of policies' performance on MuJoCo benchmark tasks under various hyper-parameter settings. For each plot, we randomly generate $30$ hyper-parameters for the policy and train for a fixed number of time steps. Reacher for $10^6$ steps, Hopper and HalfCheetah for $2\cdot10^6$ steps and SimpleHumanoid for $\approx 5\cdot10^6$ steps. The NF policy is in general more robust than Gaussian policy.}}
\label{figure:hyperparams}
\end{figure}
 
\begin{figure}[h]
\centering
\subfigure[\textbf{Reacher}: $K$]{\includegraphics[width=.23\linewidth]{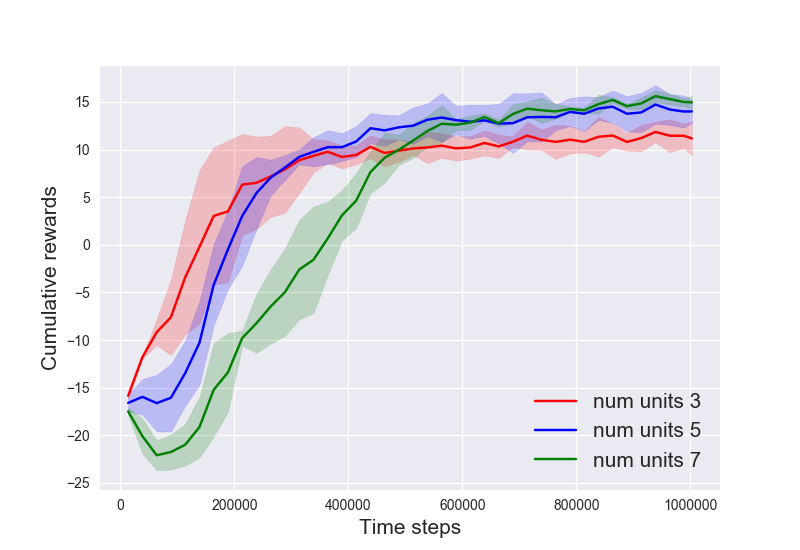}}
\subfigure[\textbf{Hopper}: $K$]{\includegraphics[width=.23\linewidth]{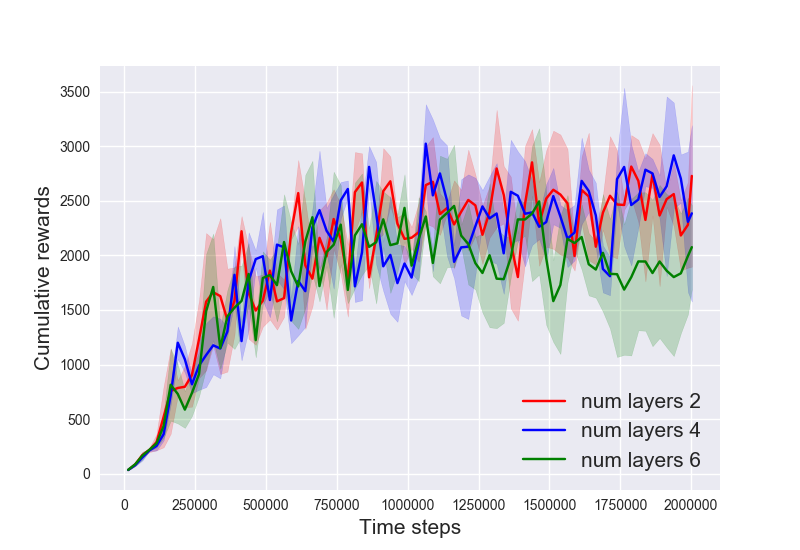}}
\subfigure[\textbf{Reacher}: $l_1$]{\includegraphics[width=.23\linewidth]{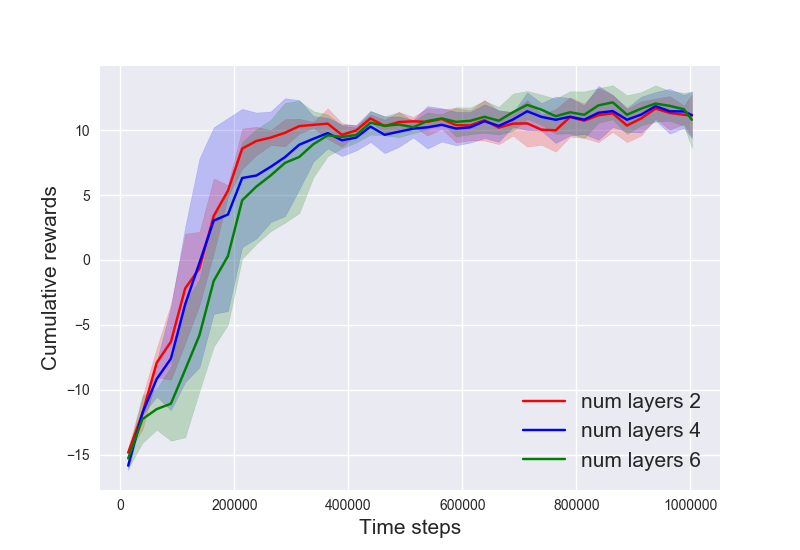}}
\subfigure[\textbf{Hopper}: $l_1$]{\includegraphics[width=.23\linewidth]{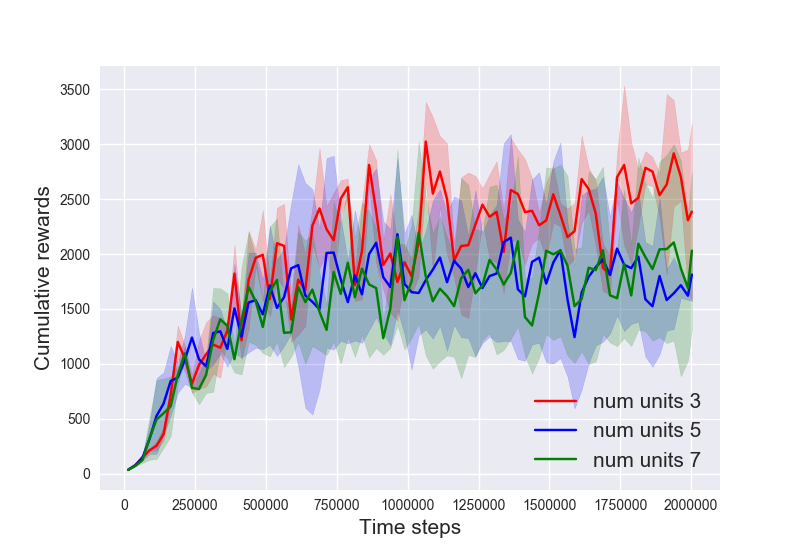}}
\caption{\small{Sensitivity to normalizing flows Hyper-parameters: training curves of the NF policy under different hyper-parameter settings (number of hidden units $l_1$ and number of transformation $K$, on Reacher and Hopper task. Each training curve is averaged over 5 random seeds and we show the mean $\pm$ std performance. Vertical axis is the cumulative rewards and horizontal axis is the number of time steps.)}}
\label{figure:ablation}
\end{figure}

\subsection{Comparison with Gaussian Policy with Big Networks}
In our implementation, the NF policy has more parameters than Gaussian policy. A natural question is whether the gains in policy optimization is due to a bigger network. To test this, we train Gaussian policy with large networks: 2-layer neural network with $128$ hidden units per layer. In Table \ref{table:networksize}, we find that for Gaussian policy, bigger network does not perform as well as the smaller network  ($32$ hidden units per layer). Since Gaussian policy with bigger network has more parameters than NF policy, this validates the claim that the performance gains of NF policy are not (only) due to increased parameters.

\begin{table}[t]
\caption{Comparison of Gaussian policy with networks of different sizes. Big network has $128$ hidden units per layer while small network has $32$ hidden units per layer. Both networks have $2$ layers. Small networks generally performs better than big networks. Below we show averageg $\pm$ std cumulative rewards after training for $5 \cdot 10^6$ steps.}
\label{policy}
\vskip 0.15in
\begin{center}
\begin{small}
\begin{sc}
\begin{tabular}{lcc}
\toprule
Task &  Gaussian (big) & Gaussian (small)\\
\midrule
Ant & $-104 \pm 30$ & $-94 \pm 44$ \\
Sim. Human. (L) & $5.1 \pm 0.7$ & $6.4 \pm 0.4$ \\
Humanoid & $501 \pm 14$ & $708  \pm 43$ \\
Humanoid (L) & $20 \pm 2$ & $53 \pm 9$ \\
\bottomrule
\end{tabular}
\end{sc}
\end{small}
\end{center}
\vskip -0.1in
\label{table:networksize}
\end{table}

\subsection{Additional Comparison with Prior works}
To ensure that we compare with state-of-the-art results of baseline TRPO, we also compare with prior works that report TRPO results.

In Table \ref{table:harrnoja2018}, we compare with Gaussian policy $+$ TRPO reported in \citep{tuomas2018}. In \citep{tuomas2018}, policies are trained for sufficient steps before evaluation, and we estimate their final performance directly from plots in \citep{tuomas2018}. For fair comparison, we record the performance of NF policy $+$ TRPO. We find that NF policy significantly outperforms the results in \citep{tuomas2018} on most reported tasks.

\paragraph{Additional Comparison with Beta Distribution.} \citet{chou2017improving} show the performance gains of Beta policy for a limited number of benchmark tasks, most of which are relatively simple (with low dimensional observation space and action space). However, they show performance gains on Humanoid-v1. We compare the results of our Figure \ref{figure:benchmarkdist} with Figure 5(j) in \citep{chou2017improving} (assuming each training epoch takes $\approx 2000$ steps): within 10M training steps, NF policy achieves 
faster progress, reaching $\approx 4000$ at the end of training while Beta policy achieves $\approx 1000$. According to \citep{chou2017improving}, Beta distribution can have an asymptotically better performance with $\approx 6000$, while we find that NF policy achieves asymptotically $\approx 5000$.

\begin{table*}[t]
\caption{Comparison with TRPO results reported in \citep{tuomas2018}. \citet{tuomas2018} train on different tasks for different time steps, we compare the performance of our method with their results with the same training steps. Better results are highlighted in bold font. When results are not statistically different, both are highlighted.}
\label{policy}
\vskip 0.15in
\begin{center}
\begin{small}
\begin{sc}
\begin{tabular}{lcccc}
\toprule
Task & NF (ours) & Gaussian (ours) & Harrnoja et al. 2018\\
\midrule
Hopper (2M) & $\mathbf{\approx 2000}$ & $\mathbf{\approx 2000}$ & $\approx 1300$ \\
HalfCheetah (10M) & $\mathbf{\approx 2900}$ & $\approx 1600$ & $\approx 1500$ \\
Walker2d (5M) & $\mathbf{\approx 3200}$ & $\approx 3000$ & $\approx 800$   \\
Ant (10M) & $\mathbf{\approx 2000}$ & $\approx 0$ & $\approx 1250$ \\
HumanoidRllab (7M) & $\mathbf{\approx 90}$ & $\approx 60$ & $\mathbf{\approx 90}$ \\
\bottomrule
\end{tabular}
\end{sc}
\end{small}
\end{center}
\vskip -0.1in
\label{table:harrnoja2018}
\end{table*}

\section{Reward Structure of Ant Locomotion Task}
For Ant locomotion task \citep{brockman2016}, the state space $\mathcal{S} \subset \mathbb{R}^{116}$ and action space $\mathcal{A} \subset \mathbb{R}^8$. The state space consists of all the joint positions and joint velocities of the Ant robot, while the action space consists of the torques applied to joints. The reward function at time $t$ is $r_t \propto v_x$ where $v_x$ is the center-of-mass velocity of along the x-axis.  In practice the reward function also includes terms that discourage large torques and encourages the Ant robot to stay alive (as defined by not flipping itself over).

Intuitively, the reward function encourages the robot to move along x-axis as fast as possible. This is reflected in Figure \ref{figure:expressive} (c) as the trajectories (red) generated by the NF policy is spreading along the x-axis. Occasionally, the robot also moves in the opposite direction.

\section{Justification from \emph{ Control as Inference} Perspective}
We justify the use of an expressive policy class from an inference perspective of reinforcement learning and control \citep{levine2018reinforcement}. The general idea is that since reinforcement learning can be cast into a structured variational inference problem, where the variational distribution is defined through the policy. When we have a more expressive policy class (e.g. Normalizinng flows) we could obtain a more expressive variational distribution, which can approximate the  posterior distribution more accurately, as observed from prior works in variational inference \citep{rezende2015}.

Here we briefly introduce the control as inference framework. Consider MDP with horizon $T$ and let $\tau = \{s_t,a_t\}_{t=0}^{T-1}$ denote a trajectory consisting of $T$ state-action pairs. Let reward $r_t = r(s_t,a_t)$ and define optimality variable as having a Bernoulli distribution $p(O_t=1|s_t,a_t) = \exp(\frac{r_t}{c})$ for some $c > 0$. We can always scale $r_t$ such that the probability is well defined. The graphical model is completed by the conditional distribution $p(s_{t+1}|s_t,a_t)$ defined by the dynamics and prior $p(a_t)$ defined as a uniform distribution over action space $\mathcal{A}$. In this graphical model, we consider $O = \{O_t\}_{t=0}^{T-1}$ as the observed variables and $\tau$ as hidden variables. The inference problem is to infer the posterior distribution $p(\tau|O)$.

If we posit the variational distribution as $q(\tau) = \Pi_{t=0}^{T-1} p(s_{t+1}|s_t,a_t) \pi(a_t|s_t)$ and optimize the variational inference objective $\min_{\pi} \mathbb{KL}[q(\tau) || p(\tau|O)]$. Under deterministic dynamics (or very good approximation when the entropy induced by stochastic dynamics is much smaller than that induced by the stochastic policy), this objective is equivalent to $\max_{\pi} \mathbb{E}_\pi [R(\tau) + c \mathbb{H}(\tau)]$, where $R(\tau) = \sum_{t=0}^{T-1} r_t$ is the total reward for trajectory $\tau$. This second objective is the maximum entropy reinforcement learning problem \cite{tuomas2017}.

The effectiveness of variational inference is largely limited by the variational distribution, which must balance optimization tractability and expressiveness. When the optimization procedure is tractable, having an expressive variational distribution usually leads to better approximation to the posterior \citep{rezende2015}. In the context of RL as inference, by construction the variational distribution $q(\tau)$ is built upon the dynamics $p(s_{t+1}|s_t,a_t)$ (which we cannot control) and policy $\pi(a_t|s_t)$ (which we are allowed to choose). By leveraging expressive policy, we can potentially benefit from a more expressive induced variational distribution $q(\tau)$, thereby improving the policy optimization procedure.

\section{Illustration of Locomotion tasks}

\begin{figure*}[h]
\centering
\includegraphics[width=.8\linewidth]{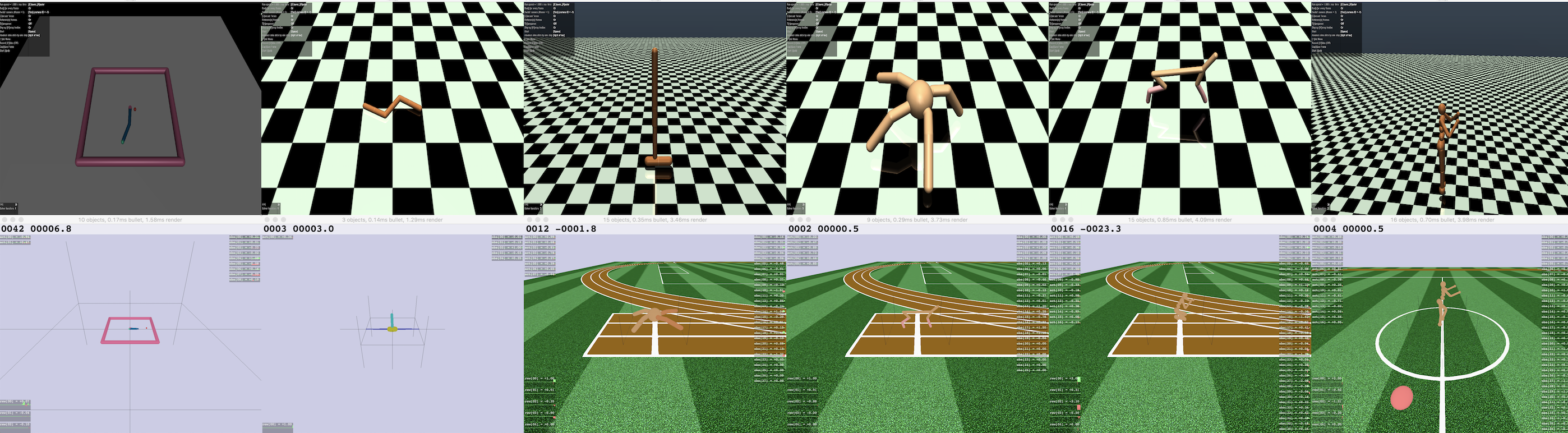}
\caption{\small{Illustration of benchmark tasks in OpenAI MuJoCo \citep{brockman2016,todorov2008}, rllab (top row) \citep{duanxi2016} and Roboschool (bottom row) \citep{schulman2017proximal}.}}
\label{figure:mujoco}
\end{figure*}

\end{document}